%% file: camera-ready.tex
\documentclass[runningheads]{llncs}

\usepackage{eccv}

\usepackage{eccvabbrv}

\usepackage{graphicx}
\usepackage{booktabs}
\usepackage{pifont}

\usepackage{enumitem}
\setlist[itemize]{label=\textbullet}
\usepackage{wrapfig}
\usepackage{multirow}
\usepackage{colortbl}
\usepackage{comment}
\usepackage{marvosym}
\usepackage{footmisc}

\usepackage[accsupp]{axessibility}  

\usepackage{hyperref}

\usepackage{orcidlink}

\begin{document}






\title{VLZip: Unified Visual and Textual Compression for Interleaved Long-Context Modeling}
\titlerunning{VLZip: Unified Visual and Textual Compression}

\author{Yuqi Zhang\inst{1,2,3}$^{\star}$\orcidlink{0009-0005-8059-5976} \and
Cheng Chen\inst{3}$^{\star}$\orcidlink{0000-0003-3662-0263} \and
Yuyu Guo\inst{3}\orcidlink{0000-0003-4376-6922} \and
Wenjie Yang\inst{3}\orcidlink{0009-0005-4773-783X} \and
\\
Lingchen Meng\inst{1}\orcidlink{0009-0006-9380-7111} \and
Peng Di\inst{3} \and
Hang Yu\inst{3}(\Letter)\orcidlink{0000-0002-5639-0912} \and
Zuxuan Wu\inst{1,2}(\Letter)\orcidlink{0000-0002-8689-5807} \and
\\
Yu-Gang Jiang\inst{1}(\Letter)\orcidlink{0000-0002-1907-8567}}
\authorrunning{Y.~Zhang et al.}

\institute{Shanghai Key Lab of Intell. Info. Processing, School of CS, Fudan University, China \and
Shanghai Innovation Institute, China \and
Ant Group, China \\
\email{zxwu@fudan.edu.cn, hyu1@e.ntu.edu.sg}}
\maketitle
\let\oldthefootnote\thefootnote
\renewcommand{\thefootnote}{}
\footnotetext{$\star$ Equal contribution. This work was done when Yuqi Zhang was a research intern at Ant Group.}
\renewcommand{\thefootnote}{\oldthefootnote}

\setcounter{footnote}{0}

\begin{abstract}
Vision Language Models (VLMs) face significant challenges with ultra-long, interleaved image-text sequences due to the quadratic complexity of self-attention. 
Current solutions either resort to aggressive token pruning, risking irreversible information loss, or adopt efficient but less precise architectures, while largely ignoring the equally vital textual component. 
We introduce VLZip, a framework that unifies visual and textual compression for high-fidelity reasoning within a pure Transformer. 
At its core, VLZip hierarchically distills visual and textual segments into compact, layer-specific "soft prefixes" and injects them into each decoder layer's hidden states, drastically shortening the attention sequence while preserving fine-grained global context. 
To address deficient evaluations in the field, we also introduce LongVLBench, a new benchmark derived from video narratives that demands holistic, narrative-level reasoning. 
Extensive experiments show VLZip achieves leading performance on long-context multimodal reasoning, enabling training up to 120K tokens—a 6$\times$ increase over the baseline—and inference beyond 280K tokens with significantly reduced memory, while demonstrating the memory scalability to handle up to 2M tokens. By excelling at extreme context lengths where existing methods collapse, VLZip establishes an efficient and powerful new standard for long-context multimodal AI. 
Code is available at \url{https://github.com/ShareLab-SII/VLZip}.
\end{abstract}

\section{Introduction}
\label{sec:intro}

The celebrated success of VLMs~\cite{DBLP:conf/nips/AlayracDLMBHLMM22, DBLP:conf/icml/0008LSH23, DBLP:conf/nips/LiuLWL23a, DBLP:conf/cvpr/LiuLLL24} has been largely confined to a world of snapshots, not narratives. While highly proficient at processing single images, they struggle with the ultra-long, interleaved sequences of images and text that define complex real-world activities. These applications span a wide spectrum—from following detailed illustrated manuals and analyzing a patient's comprehensive visual medical history, to evaluating long-form video narratives and powering multimodal GUI agents that autonomously operate over extended horizons. Across all these domains, the next frontier for AI is not merely to process more tokens, but to unlock a new class of \textbf{long-context intelligence} capable of holistic, narrative-level understanding.

However, this critical domain remains largely underexplored. 
Progress is stalled by two fundamental, interconnected hurdles: the first is \textbf{architectural}, with the quadratic scaling of self-attention~\cite{DBLP:conf/nips/VaswaniSPUJGKP17} making long sequences computationally prohibitive; the second is \textbf{evaluative}: the benchmarks used to measure progress are poorly suited to narrative reasoning\footnote{Throughout this paper, we use the term \textit{reasoning} to refer to the model's broad capacity for interpreting and synthesizing multimodal information, as distinct from explicit chain-of-thought or deliberative "thinking" paradigms.}, which steers the research community away from solving the real problem. 

\begin{wrapfigure}{r}{0.48\linewidth}  
    \centering
    \includegraphics[width=0.46\textwidth]{{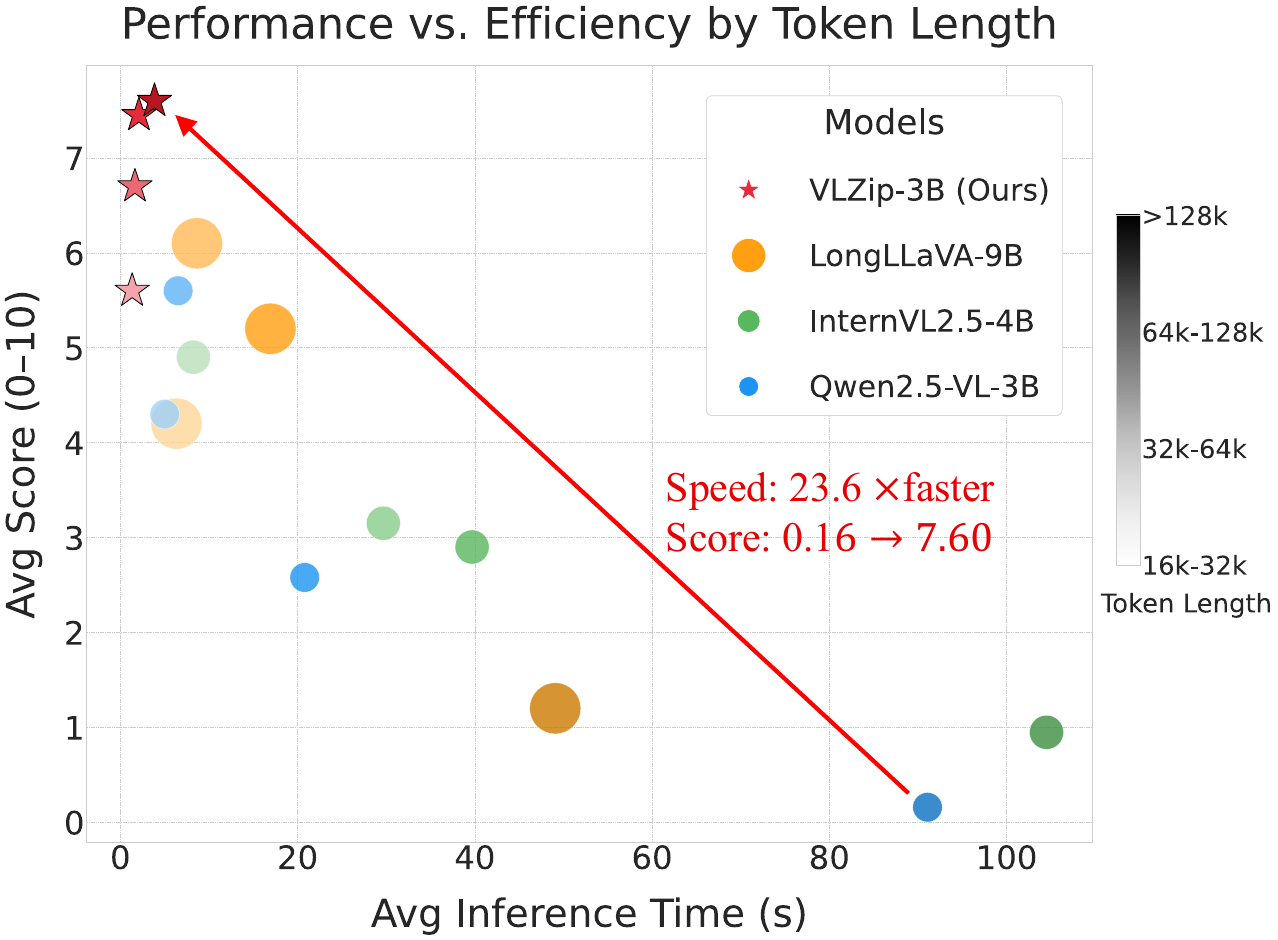}}
    
    \caption{A comparison of performance and efficiency for VLZip(Ours, red star) and baseline models. 
Performance across context lengths is indicated by the color intensity of the points.
}
    \label{fig:performance_vs_efficiency}
\end{wrapfigure}

On the architectural side, existing approaches typically address this challenge by accepting a trade-off between efficiency and fidelity (as shown in Fig.~\ref{fig:performance_vs_efficiency}). One path involves aggressive input pruning~\cite{DBLP:journals/corr/abs-2508-01548}, a strategy that irreversibly discards tokens—almost exclusively from the visual modality—risking significant information loss. A second path attempts an architectural overhaul, replacing the Transformer with more efficient but potentially less expressive backbones like State Space Models~\cite{DBLP:journals/corr/abs-2409-02889,DBLP:conf/iclr/YeXF0HYDK0L25}, which can compromise the precise reasoning at which Transformers excel. A third, data-centric strategy~\cite{DBLP:journals/tmlr/JiangHZWKLC24} improves reasoning through high-quality data but fails to address the computational bottleneck. 
Yet none offers a principled path to simultaneously high efficiency and high-fidelity reasoning within a pure Transformer.

On the evaluation side, the few existing long-context multimodal benchmarks are equally unsatisfying. Some, such as Mantis-Eval~\cite{DBLP:journals/tmlr/JiangHZWKLC24} and MileBench~\cite{DBLP:journals/corr/abs-2404-18532}, are built by repurposing short-context datasets, leading to imbalanced image-to-text ratios or insufficient context lengths. Others rely on artificial "needle-in-a-haystack" (NIAH) retrieval tasks, as seen in MM-NIAH~\cite{DBLP:conf/nips/WangZRDLLH0ZLZL24} and MMLongCite~\cite{DBLP:journals/corr/abs-2510-13276}, which test for simple fact recall rather than holistic comprehension. MMLongBench~\cite{wang2025mmlongbenchbenchmarkinglongcontextvisionlanguage}, for instance, creates length by padding with irrelevant filler documents, introducing noise and risking data contamination. This creates a self-perpetuating limitation: models are not designed for genuine narrative intelligence because the benchmarks do not demand it.

We argue that true progress requires a unified solution that addresses both modeling and evaluation. In this paper, we introduce VLZip, a framework that redefines the efficiency-fidelity curve by introducing a more intelligent trade-off. Instead of sacrificing reasoning or irreversibly discarding information, VLZip trades a small, one-time compression cost for massive, sustained efficiency gains within a pure Transformer architecture. At its core, VLZip operates on a principle of hierarchical context distillation and multi-layer injection. It first distills extensive image and text segments into compact, information-rich "soft prefixes", and then injects these prefixes into the hidden states of every decoder layer. This provides the model with a constant, fine-grained stream of global context, allowing it to maintain a high-fidelity understanding of the entire narrative while operating on a drastically shortened sequence. As a preview of our results, Fig.~\ref{fig:performance_vs_efficiency} illustrates how VLZip substantially improves the existing performance-efficiency trade-off, achieving state-of-the-art accuracy while being over an order of magnitude faster than baselines on ultra-long contexts.

To properly validate VLZip and address the field's evaluation gap, we also construct a new, challenging benchmark, \textbf{LongVLBench}, derived from video narratives. This benchmark provides long, chronologically coherent sequences that force models to move beyond simple retrieval and engage in authentic narrative-level reasoning. Our extensive experiments demonstrate that VLZip achieves strong performance on MMLongBench—particularly excelling on in-context learning tasks at extreme lengths—and sets a new state-of-the-art on our narrative-focused LongVLBench where other methods falter. It enables training on sequences up to 120K tokens on standard hardware—a 6$\times$ improvement over the baseline—while drastically reducing computational and memory costs.

In summary, the contributions of this paper are:
\begin{itemize} [leftmargin=*,itemsep=0.2pt,topsep=0pt,partopsep=0pt]
    \item We introduce VLZip, a unified visual-textual compression framework that enables training on sequences up to 120K tokens and inference beyond 280K tokens, while demonstrating a memory-efficient design that shows a clear path to scaling to 2M tokens within a pure Transformer architecture.
    \item We propose hierarchical context distillation with multi-layer injection that, for the first time, unifies the compression of both visual and textual modalities into layer-specific "soft prefixes" injected into every decoder layer.
    \item We construct a benchmark, LongVLBench, for evaluating long-context narrative understanding. Derived from video, it provides chronologically coherent interleaved sequences that address the shortcomings of existing benchmarks (e.g., artificial tasks, imbalanced contexts) and require narrative reasoning.
    \item We empirically demonstrate that VLZip both sets a new state-of-the-art (SOTA) on long-context multimodal reasoning benchmarks and enables training on sequences up to 120K tokens on standard hardware—a 6-fold improvement over the baseline—establishing a practical standard for long-context AI.
\end{itemize}

\section{Related Works}
\label{sec:related_works}
The pursuit of long-context VLMs is an emerging field, with progress confronting challenges in both model architecture and evaluation. A detailed review is provided in the supplementary material; here, we contextualize our work by briefly surveying the dominant strategies.

\noindent\textbf{Architectures.} Current long-context VLM architectures have largely centered on three strategies, each forcing a compromise between performance and efficiency. The first, Input Pruning~\cite{DBLP:journals/corr/abs-2508-01548,DBLP:conf/cvpr/YangCTWL0J25}, permanently discards visual tokens to shorten sequences, risking irreversible information loss while ignoring the equally vast textual information present in interleaved multimodal sequences. A second path, Architectural Overhaul~\cite{DBLP:journals/corr/abs-2409-02889,DBLP:conf/iclr/YeXF0HYDK0L25}, replaces the Transformer's attention mechanism with more scalable backbones like State Space Models, but can sacrifice the precise, non-local reasoning capabilities at which Transformers excel. A third, Data-Centric approach~\cite{DBLP:journals/tmlr/JiangHZWKLC24} enhances multi-image reasoning with high-quality data but does not resolve the computational bottleneck for long sequences.
Our work is inspired by a promising alternative from the NLP domain: Soft Prompt Compression~\cite{cheng2024xrag,liao2025e2llm,chevalier2023adapting,ge2024context}. Methods such as GIST~\cite{mu2023learning} and the In-Context Autoencoder~\cite{ge2024context} demonstrate that long textual contexts can be effectively distilled into compact, information-rich soft prompts, offering efficiency without aggressive pruning. However, their application has been confined to the text modality and typically involves a simple, single-layer injection. Concurrently, DeepStack~\cite{meng2024deepstack} explores multi-layer stacking for visual tokens but does not address textual compression or interleaved multimodal contexts. VLZip bridges these two lines of work by unifying both visual and textual compression with multi-layer injection in a single framework, preserving high-fidelity context throughout decoding.

\noindent\textbf{Evaluations.}
Paralleling the architectural challenges, existing multimodal long-context benchmarks also exhibit shortcomings. Many are constructed by repurposing short-context datasets~\cite{DBLP:journals/tmlr/JiangHZWKLC24,DBLP:journals/corr/abs-2404-18532}, relying on artificial `needle-in-a-haystack' retrieval tasks~\cite{DBLP:conf/nips/WangZRDLLH0ZLZL24,DBLP:journals/corr/abs-2510-13276}, or padding with irrelevant filler documents~\cite{wang2025mmlongbenchbenchmarkinglongcontextvisionlanguage}. These limitations motivate our development of LongVLBench, a benchmark built from the ground up with chronologically coherent, narrative-driven sequences to enable a more authentic evaluation of holistic reasoning.

\section{Method}

We introduce VLZip, a unified compression framework for efficient, high-fidelity reasoning over long interleaved image-text sequences. A complete input sequence, formally expressed as $\mathcal{X} = \{\mathcal{S}, \mathcal{C}, \mathcal{Q} \}$, comprises a system prompt $\mathcal{S}$, a question $\mathcal{Q}$, and the lengthy interleaved context $\mathcal{C}$. 
To reduce this cost, VLZip compresses only the context $\mathcal{C}$, leaving the prompt and question untouched. Instead of feeding the full context to the VLM, it represents each image and text segment with compact placeholder tokens, drastically shortening the effective input length.

The interleaved context $\mathcal{C} = \{t_0, \mathbf{v}_0, \ldots, t_{n-1}, \mathbf{v}_{n-1},$ $t_n\}$, where $t_i$ are text segments and $\mathbf{v}_j$ are images, is processed externally by modality-specific compressors. 
These hierarchical compressors distill each image and text segment into compact, layer-specific soft-prefix features. Rather than being appended to the input as additional tokens, these features are injected at the positions held by placeholder tokens during decoding. 
Critically, at each decoder layer, the corresponding compressed features are injected into the hidden states of the placeholders, ensuring continuous, fine-grained access to global context and overcoming information loss from single-step pruning or shallow injection.

\subsection{Architecture}

\begin{figure*}[t]
\centering
\includegraphics[width=1\linewidth]{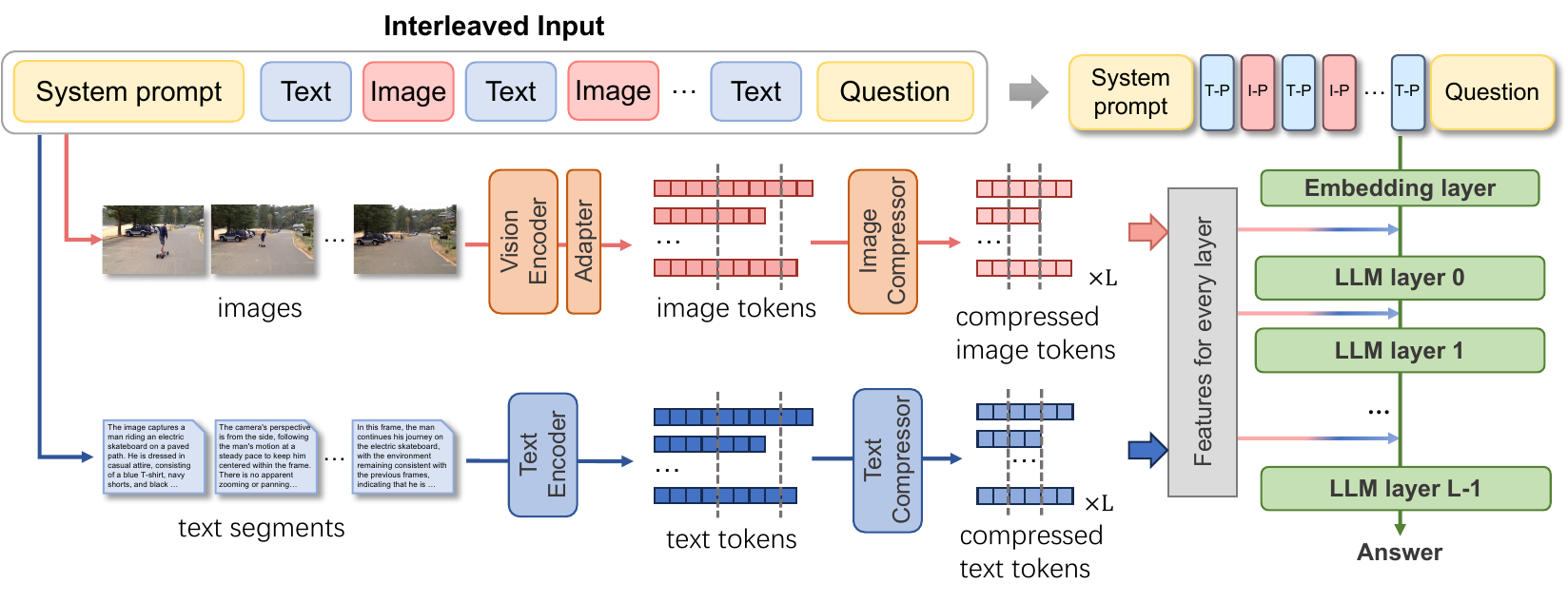}
\caption{\textbf{Overview of VLZip Architecture.} 
VLZip compresses interleaved image-text sequences via modality-specific hierarchical modules.   
The visual module partitions each image into chunks and compresses each into $M_v$ layer-specific tokens via a shared Q-Former. The textual module partitions each text segment into chunks, encodes each chunk with a lightweight text encoder, and compresses each into $M_t$ layer-specific tokens via a shared Q-Former.
Compressed features are injected into each LLM decoder layer via element-wise addition.
T-P and I-P denote text and image placeholders.}
\label{fig:Architecture}
\end{figure*}

As illustrated in Figure~\ref{fig:Architecture}, the VLZip framework is built upon a standard VLM backbone and comprises three key components: (1) a \textbf{Hierarchical Visual Compressor} that encodes each image into a set of layer-specific feature vectors; (2) a \textbf{Hierarchical Textual Compressor} that similarly distills long text segments into layer-specific features; and (3) an \textbf{Information Injection Mechanism} that injects the compressed features into dedicated placeholder tokens at every decoder layer.
The placeholder tokens preserve the compact sequence
structure, while the layer-specific soft-prefix features
carry the semantic information of the original context.

\noindent\textbf{Hierarchical Visual Compression Module.}
In a conventional VLM, each image $v_j$ is passed through a vision encoder and a projection adapter to produce a sequence of visual tokens $\mathbf{F}_{v,j} \in \mathbb{R}^{N_{v,j} \times d}$, where $d$ is the VLM's hidden dimension and the token count $N_{v,j}$ can be hundreds or thousands. To avoid this bottleneck, we adaptively compress these $N_{v,j}$ tokens into a smaller set of $N_{v,j}'$ tokens.

Specifically, each image is uniformly partitioned into chunks of size $C_v$, with each independently processed by a shared image compressor that generates $M_v$ compressed visual tokens per decoder layer. The total number of compressed tokens per image is thus $N_{v,j}' = \lceil N_{v,j} / C_v \rceil \times M_v$.

To avoid irreversible information loss, the image compressor employs a hierarchical strategy by producing layer-specific compressed features for all $L$ decoder layers simultaneously. It utilizes a Q-Former architecture with a shared set of $L \times M_v$ learnable queries $\mathbf{q}_v \in \mathbb{R}^{(L \cdot M_v) \times d}$, allowing all layer-specific tokens to be generated in a single forward pass. For a given image chunk $\mathbf{F}_{v,j,p} \in \mathbb{R}^{C_v \times d}$ (the $p$-th chunk of image $v_j$), the compression process is:
\begin{align}
\mathbf{Q}_v = \text{LN}(\mathbf{q}_v); \quad\mathbf{K}_v &= \text{LN}(\mathbf{F}_{v,j,p} + \text{PE}(\mathbf{F}_{v,j,p})); \quad \mathbf{V}_v = \text{LN}(\mathbf{F}_{v,j,p}) \\
\mathbf{Z}_{v,j,p} &= \text{MHA}(\mathbf{Q}_v, \mathbf{K}_v, \mathbf{V}_v)
\end{align}
Here, $\text{LN}(\cdot)$ denotes Layer Normalization, $\text{PE}(\cdot)$ denotes Positional Encoding, and $\text{MHA}(\cdot)$ denotes Multi-Head Attention. The output $\mathbf{Z}_{v,j,p} \in \mathbb{R}^{(L \cdot M_v) \times d}$ is reshaped into $\mathbb{R}^{L \times M_v \times d}$, where the $l$-th slice $\mathbf{Z}_{v,j,p}^{(l)} \in \mathbb{R}^{M_v \times d}$ gives the $M_v$ compressed visual tokens for chunk $p$ tailored for decoder layer $l$.

\noindent\textbf{Hierarchical Text Compression Module.}
Long text segments present a similar computational burden. For a given text segment $t_i$ with $N_{t,i}$ tokens, we first uniformly partition it into $K_i = \lceil N_{t,i} / C_t \rceil$ chunks of size $C_t$. Each chunk is then independently encoded by a lightweight text encoder $E(\cdot)$ to obtain chunk-level hidden states $\mathbf{F}_{t,i,k} = E(t_{i,k}) \in \mathbb{R}^{C_t \times d_e}$. Each chunk is then processed by the text compressor, which projects it to the VLM's hidden dimension $d$ and compresses it into $M_t$ tokens for each of the $L$ decoder layers.
The text compressor utilizes a Q-Former architecture with a shared set of $L \times M_t$ learnable queries $\mathbf{q}_t \in \mathbb{R}^{(L \cdot M_t) \times d}$ to generate all layer-specific tokens in a single forward pass. The compression of a single chunk $\mathbf{F}_{t,i,k}$ is:
\begin{align}
\mathbf{F}_{t,i,k}' &= \mathbf{F}_{t,i,k} \mathbf{W}_{\text{proj}} \\
\mathbf{Q}_t = \text{LN}(\mathbf{q}_t);\quad\mathbf{K}_t &= \text{LN}(\mathbf{F}_{t,i,k}' + \text{PE}(\mathbf{F}_{t,i,k}')) ; \quad \mathbf{V}_t = \text{LN}(\mathbf{F}_{t,i,k}') \\
\mathbf{Z}_{t,i,k} &= \text{MHA}(\mathbf{Q}_t, \mathbf{K}_t, \mathbf{V}_t)
\end{align}
where $\mathbf{W}_{\text{proj}} \in \mathbb{R}^{d_e \times d}$ is a projection matrix mapping from the text encoder's hidden dimension $d_e$ to the VLM's hidden dimension $d$. The output $\mathbf{Z}_{t,i,k} \in \mathbb{R}^{(L \cdot M_t) \times d}$ is reshaped into $\mathbb{R}^{L \times M_t \times d}$, where the $l$-th slice $\mathbf{Z}_{t,i,k}^{(l)} \in \mathbb{R}^{M_t \times d}$ gives the $M_t$ compressed text tokens for chunk $k$ tailored for decoder layer $l$.

\noindent\textbf{Information Injection.}
The VLM decoder operates on a compact sequence consisting of the system prompt $\mathcal{S}$, the question $\mathcal{Q}$, and a placeholder sequence $\mathcal{P}$ representing the compressed image and text segments. For each image chunk $\mathbf{F}_{v,j,p}$, $\mathcal{P}$ contains $M_v$ visual placeholders, while for each text chunk $\mathbf{F}_{t,i,k}$, it contains $M_t$ textual placeholders. At decoder layer $l$, the hidden states of these placeholders are augmented with the corresponding compressed features before self-attention.
Let $\mathbf{H}_{v,j,p}^{(l)} \in \mathbb{R}^{M_v \times d}$ and $\mathbf{H}_{t,i,k}^{(l)} \in \mathbb{R}^{M_t \times d}$ be the hidden states of the placeholders for the $p$-th chunk of image $v_j$ and the $k$-th chunk of text segment $t_i$, respectively. The injection is:
\begin{equation}
\hat{\mathbf{H}}_{v,j,p}^{(l)} = \mathbf{H}_{v,j,p}^{(l)} + \mathbf{Z}_{v,j,p}^{(l)}, \quad \hat{\mathbf{H}}_{t,i,k}^{(l)} = \mathbf{H}_{t,i,k}^{(l)} + \mathbf{Z}_{t,i,k}^{(l)}
\end{equation}

This element-wise addition fuses the distilled global context directly into the model's processing pipeline at every layer. The full sequence of updated hidden states is then fed into the layer's self-attention module. This allows the VLM to reason with a complete understanding of the long context while the attention mechanism operates only on a short, computationally feasible sequence.

\subsection{Training Pipeline}

To effectively train the various components of VLZip, we design a four-stage progressive training pipeline. This curriculum enables the model to first master modality-specific compression and then learn to integrate these compressed representations for complex, long-context reasoning. 

\noindent\textbf{Stage 1: Visual Compressor Pre-training.}
We train the visual compressor on large-scale single-image instruction data (e.g., LLaVA-OneVision~\cite{DBLP:journals/tmlr/0080ZGZ00ZZL0L25_llavaov} single-image data). 
Only the projection adapter and vision Q-Former parameters are trainable.
This stage teaches the compressor to distill visual tokens into compact, layer-specific representations $\mathbf{Z}_{v,j}^{(l)}$.

\noindent\textbf{Stage 2: Textual Compressor Pre-training.}
We pre-train the textual compressor using a reconstruction task on text-only long-document corpora (e.g., ChatQA2~\cite{xu2024chatqa}). The model learns to compress each text chunk into tokens $\mathbf{Z}_{t,i,k}$ and reconstruct the original text. Only the text compressor parameters (the lightweight encoder, $\mathbf{W}_{\text{proj}}$, and text Q-Former) are trained. This self-supervised objective ensures the compressed tokens preserve semantic completeness.

\noindent\textbf{Stage 3: Joint Interleaved Fine-tuning.}
With both compressors pre-trained, this stage aims to teach the VLM how to fuse and reason over compressed information from both modalities simultaneously. We use datasets containing interleaved sequences of multiple images and texts. In this stage, we keep the compressors trainable and unfreeze the LLM decoder. 
This step is crucial for harmonizing the independently trained modules within the unified architecture.

\noindent\textbf{Stage 4: Long-Context Consolidation.}
Building on previous stages, we continue training on long-form textual data with all Stage 3 components remaining trainable. This final stage adapts the model's reasoning strategies to ultra-long contexts, ensuring robust performance across extensive image-text sequences. Complete hyperparameters are provided in the supplementary material.

\section{LongVLBench: A Narrative-Driven Benchmark for Long-Context Evaluation}
\label{sec:dataset_creation}

\begin{figure*}[t]
\centering
\includegraphics[width=1\linewidth]{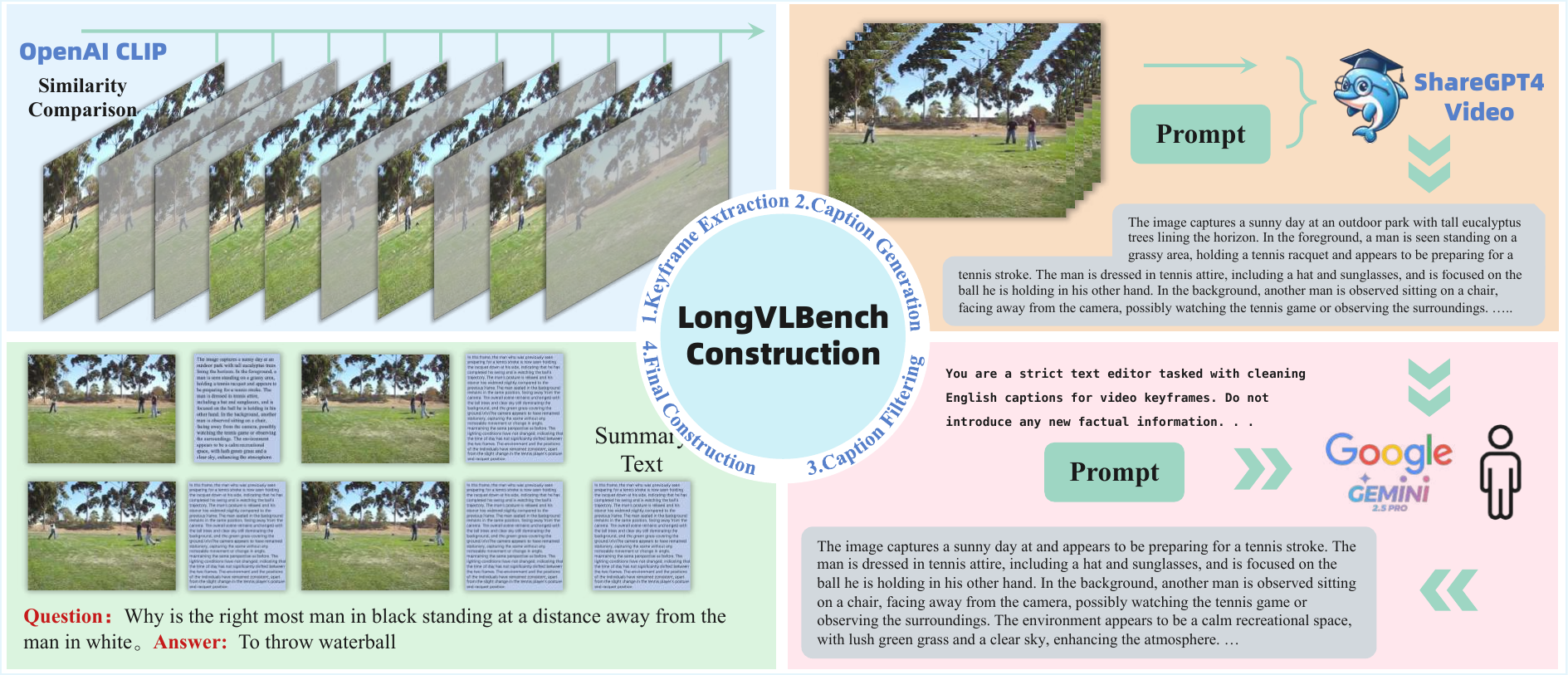}
\caption{
\textbf{Overview of our four-stage dataset creation pipeline.} 
The process begins with (1) extracting semantic keyframes from videos using CLIP-based similarity comparison. 
Next, (2) ShareGPT4Video generates detailed, transitional, and summary captions for the frames. 
These captions are then (3) rigorously refined through a prompt-guided LLM and human review to eliminate redundancy and ensure factual fidelity. 
Finally, (4) the cleaned captions and their corresponding keyframes are assembled into the final long-context, interleaved dataset.
}
\label{fig:dataset}
\end{figure*}

As discussed in Sections 1 and 2, existing long-context multimodal benchmarks are limited by several notable shortcomings. They often repurpose disconnected, short-context data, rely on artificial "needle-in-a-haystack" (NIAH) retrieval tasks that test recall over reasoning, or pad sequences with irrelevant filler documents. 
These shortcomings create a landscape where models are not evaluated on—and therefore not optimized for—genuine narrative-level understanding. To address this critical evaluation gap, we introduce \textbf{LongVLBench}, a new benchmark constructed from the ground up to provide a rigorous and authentic testbed for long-form multimodal intelligence. Sourced from video narratives, our systematic pipeline (Figure~\ref{fig:dataset}) is purpose-built to generate genuinely long, chronologically coherent, and densely interleaved sequences that demand holistic comprehension. The process consists of four key stages.

\noindent\textbf{Semantic Keyframe Extraction.}
To keep the true visual continuity, our pipeline begins by converting video into a discrete sequence of keyframes. Unlike methods that stitch together unrelated images, we employ a semantically aware extraction algorithm using CLIP embeddings~\cite{DBLP:conf/icml/RadfordKHRGASAM21}. A new keyframe is selected only when a significant visual change is detected, ensuring the resulting image sequence captures the video's core narrative arc. This process yields a chronologically coherent visual story, overcoming the limitations of benchmarks built from repurposed, disjointed data.

\noindent\textbf{Hierarchical Narrative Captioning.}
To create a context where text is an integral part of the story rather than just filler, we generate a rich, multi-layered textual narrative for the keyframes. Using a powerful video language model (ShareGPT4Video)~\cite{DBLP:conf/nips/0016WLD0ZCDB00024}, we produce three levels of description:
First, it generates a detailed caption for each individual keyframe. Then, by providing context from previous frames, it creates transitional descriptions that highlight the progression between them. Finally, a comprehensive summary caption is generated for the entire keyframe sequence to capture the video's overarching theme.

\noindent\textbf{Rigorous Refinement for Factual Fidelity.}
To ensure LongVLBench is a reliable benchmark, all auto-generated captions undergo stringent, hybrid refinement. We use a SOTA LLM (Gemini 2.5 Pro) guided by strict principles to eliminate redundancy, normalize style, and enhance fluency. Most importantly, the process is explicitly constrained to prevent the model from hallucinating facts or inferring details not present in the source material. Human review provides a final 
layer of quality control, guaranteeing that the final textual narrative is coherent, concise, and factually grounded in the visual evidence.

\noindent\textbf{Final Interleaved Data Construction.}
One crucial step is designing tasks that measure holistic reasoning, not simple fact retrieval. For each long-form document, we generate QA pairs that explicitly require synthesizing information across long spans of interleaved text and images. Questions are designed to be unanswerable by observing a single frame or reading a single caption, compelling \begin{wrapfigure}{r}{0.6\linewidth}  
    \centering
    \includegraphics[width=0.58\textwidth]{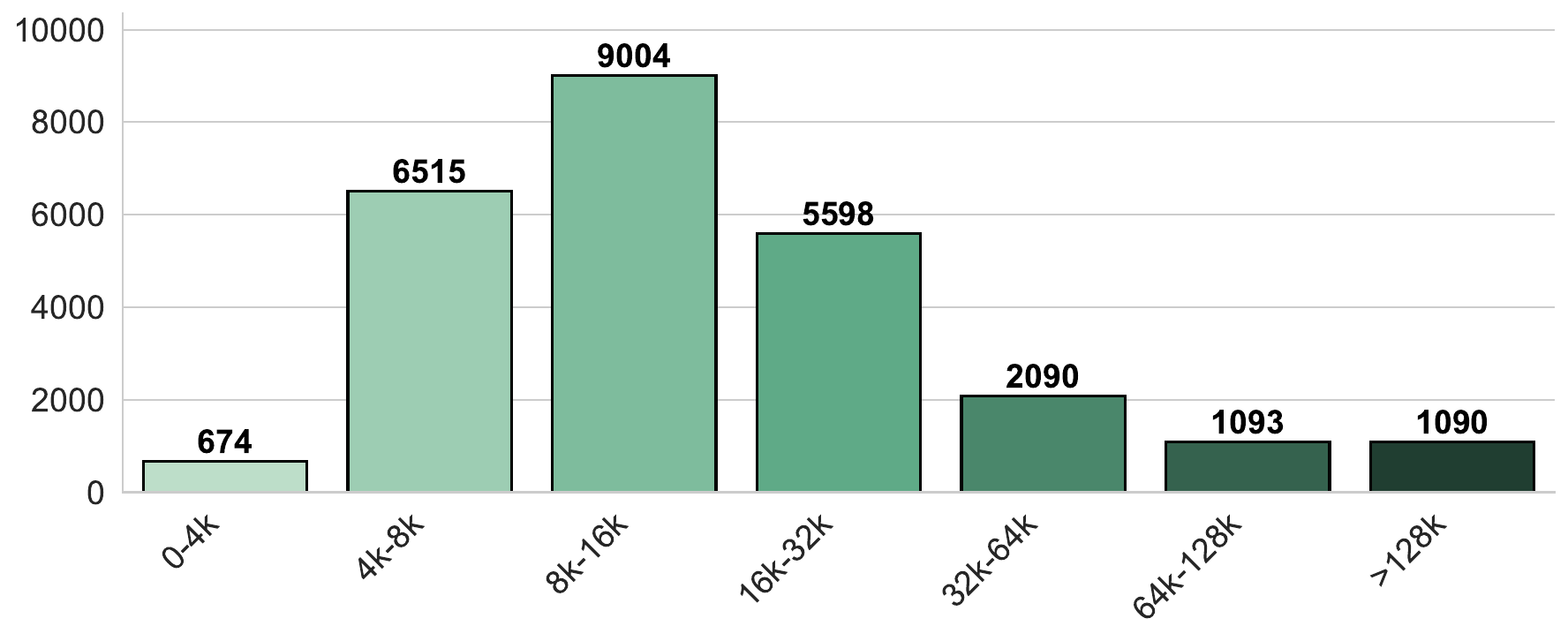}
    \caption{Token length distribution of the interleaved dataset. 
The histogram shows the frequency of the 26,164 generated documents across seven token length bins. 
}
    \label{fig:token_distri}
\end{wrapfigure}the model to understand plot progression, character interactions, and evolving context. The refined, interleaved narratives and corresponding QA pairs are assembled into final documents. From this collection, we constructed a high-quality test set of 140 samples, carefully selected to represent a diverse distribution of token lengths (Fig.~\ref{fig:token_distri}) and provide a challenging test for ultra-long multimodal reasoning.

\section{Experiments}
\subsection{Experiment Setup}
\noindent\textbf{Implementation Details.}
Our model uses the Qwen2.5-VL-Instruct-3B backbone~\cite{bai2025qwen2}. Both image and text inputs are segmented into chunks of 100 tokens, each compressed into 4 tokens per decoder layer via modality-specific Q-Former compressors. Text chunks are encoded by Qwen2.5-0.5B prior to compression. The model is trained over four stages on 32$\times$A100 or 32$\times$H20 GPUs. Additional hyperparameters are detailed in the supplementary material.

\noindent\textbf{Evaluation Benchmarks.}
We assess long-context understanding using our proposed \textbf{LongVLBench} and the established \textbf{MMLongBench}~\cite{wang2025mmlongbenchbenchmarkinglongcontextvisionlanguage}, featuring input lengths from 8K to 128K tokens. To assess foundational abilities, we also test on several short-context VQA benchmarks via the \textbf{LMMs-Eval} framework~\cite{Zhang_2025_lmms-eval}.


\noindent\textbf{Baselines.}
We compare our approach against models with distinct architectural strategies: dedicated long-context models (LongLLaVA~\cite{DBLP:journals/corr/abs-2409-02889}, Long-VITA~\cite{shen2025longvita}), interleaved reasoning models (Mantis~\cite{DBLP:journals/tmlr/JiangHZWKLC24}, InternVL2.5~\cite{chen2024expanding_internvl2_5}, Ovis2~\cite{lu2024ovis}), and other compression methods (GlimpsePrune~\cite{DBLP:journals/corr/abs-2508-01548}, VisionZip~\cite{DBLP:conf/cvpr/YangCTWL0J25}). For a fair comparison, the compression-based methods are implemented on the same Qwen2.5-VL-3B backbone. The unmodified base model is used as a reference.

\definecolor{mygreen}{rgb}{0,0.6,0}
\definecolor{myred}{rgb}{0.8,0,0}

\newcommand{\perfcell}[2]{%
  #1$_{%
    \ifnum#2=20 
      \textcolor{mygreen}{\,100\%}%
    \else 
      \textcolor{myred}{\,\the\numexpr#2*5\relax\%}%
    \fi
  }$%
}

\begin{table*}[t]
\centering
\caption{
LongVLBench performance across different context lengths. 
Colored numbers indicate the inference success rate (\textcolor{mygreen}{100\%} denotes full success, \textcolor{myred}{lower rates} indicate OOM or no responses). 
The overall average is computed as the total score divided by the total number of samples across all length ranges and scaled to 0-100 range. 
\textbf{Bold} and \underline{underline} indicate the best and second best performance respectively.
}
  \label{tab:qa_performance}
\renewcommand\arraystretch{1.2}
\renewcommand\tabcolsep{4.0pt}
\centering
\resizebox{\linewidth}{!}{
\begin{tabular}{lccccccccc}
\toprule
\multicolumn{1}{c}{\textbf{Model}} & \textbf{Size}& \textbf{0-4k} & \textbf{4k-8k} & \textbf{8k-16k} & \textbf{16k-32k} & \textbf{32k-64k} & \textbf{64k-128k} & \textbf{\textgreater 128k} & \textbf{Avg} \\
\midrule
Long-VITA~\cite{shen2025longvita} & 14B & \perfcell{\underline{5.25}}{20} & \perfcell{\underline{6.10}}{20} & \perfcell{\textbf{6.94}}{16} & \perfcell{3.80}{15} & \perfcell{5.78}{9} & \perfcell{\underline{5.33}}{3} & \perfcell{0.00}{0} & 33.1 \\
\addlinespace
LongLLaVA~\cite{DBLP:journals/corr/abs-2409-02889} & 9B & \perfcell{4.70}{20} & \perfcell{5.15}{20} & \perfcell{4.95}{20} & \perfcell{4.20}{20} & \perfcell{\underline{6.10}}{20} & \perfcell{5.20}{20} & \perfcell{1.20}{20} & 45.0 \\
\addlinespace
Mantis~\cite{DBLP:journals/tmlr/JiangHZWKLC24} & 8B & \perfcell{0.70}{20} & \perfcell{2.00}{20} & \perfcell{1.88}{17} & \perfcell{0.00}{2} & \perfcell{2.00}{1} & \perfcell{0.00}{0} & \perfcell{0.00}{0} & 6.30 \\
\addlinespace
Qwen2.5-VL~\cite{bai2025qwen2} & 7B & \perfcell{4.90}{20} & \perfcell{5.55}{20} & \perfcell{5.80}{20} & \perfcell{\underline{5.25}}{20} & \perfcell{5.70}{20} & \perfcell{4.25}{20} & \perfcell{\underline{1.58}}{19} & \underline{47.1} \\
\addlinespace
InternVL2.5~\cite{chen2024expanding_internvl2_5} & 4B & \perfcell{4.95}{20} & \perfcell{5.05}{20} & \perfcell{5.85}{20} & \perfcell{4.90}{20} & \perfcell{3.15}{20} & \perfcell{2.90}{20} & \perfcell{0.95}{19} & 39.1 \\
\addlinespace
Ovis2~\cite{lu2024ovis} & 4B & \perfcell{4.85}{20} & \perfcell{\textbf{6.25}}{20} & \perfcell{6.25}{20} & \perfcell{4.35}{20} & \perfcell{5.95}{20} & \perfcell{0.4}{20} & \perfcell{0.5}{20} & 40.8 \\
\addlinespace
GlimpsePrune~\cite{DBLP:journals/corr/abs-2508-01548} & 3B & \perfcell{\textbf{5.35}}{20} & \perfcell{4.70}{20} & \perfcell{6.20}{20} & \perfcell{4.45}{20} & \perfcell{5.10}{20} & \perfcell{2.21}{19} & \perfcell{0.00}{13} & 39.9 \\
\addlinespace
VisionZip~\cite{DBLP:conf/cvpr/YangCTWL0J25} & 3B & \perfcell{4.95}{20} & \perfcell{4.70}{20} & \perfcell{\underline{6.89}}{18} & \perfcell{3.75}{4} & \perfcell{2.00}{1} & \perfcell{4.00}{3} & \perfcell{0.00}{0} & 24.7 \\			
\addlinespace
Qwen2.5-VL~\cite{bai2025qwen2} & 3B & \perfcell{\textbf{5.35}}{20} & \perfcell{4.90}{20} & \perfcell{6.20}{20} & \perfcell{4.30}{20} & \perfcell{5.60}{20} & \perfcell{2.58}{19} & \perfcell{0.16}{19} & 41.4 \\
\addlinespace
VLZip (Ours) & 3B & \perfcell{4.60}{20} & \perfcell{5.65}{20} & \perfcell{5.75}{20} & \perfcell{\textbf{5.60}}{20} & \perfcell{\textbf{6.70}}{20} & \perfcell{\textbf{7.45}}{20} & \perfcell{\textbf{7.60}}{20} & \textbf{61.9} \\
\bottomrule
\end{tabular}}
\end{table*}

\subsection{Main Results}
\noindent\textbf{Results on LongVLBench.}
On our challenging LongVLBench, designed to test authentic narrative-level reasoning, VLZip sets a new state-of-the-art with an average score of \textbf{61.9}, a 31.4\% relative improvement over the next-best model (Tab.~\ref{tab:qa_performance}).
This success stems from our unified compression, which enables robust processing of all test samples at 100\% inference success rate across all length ranges.
Despite being 3$\times$ smaller, our 3B model consistently outperforms the 9B LongLLaVA.
The advantage is starkest at extreme lengths: while the baseline Qwen2.5-VL-3B model nearly fails on inputs exceeding 128k tokens (scoring only 0.16), VLZip maintains strong performance at 7.60, 
an absolute improvement of 7.44 points, highlighting that unified compression is critical for sustaining narrative comprehension at ultra-long scales.

\begin{table*}[t]
\centering
\caption{Comparison on VRAG, NIAH, and ICL tasks of MMLongBench. 
*Text segments shorter than 100 characters are passed through without 
compression, as such short segments yield minimal efficiency gains 
while compression may discard critical information.
 \textbf{Bold} and \underline{underline} indicate the best and second best.
}
\label{tab:model_performance_mmlongbench}

\resizebox{\textwidth}{!}{%
\begin{tabular}{l l | ccc | ccccc >{\columncolor{gray!15}}c}
\toprule
\multicolumn{2}{c|}{\textbf{Metric}} & \multicolumn{3}{c|}{\textbf{Large Scale Models (\textgreater=7B)}} & \multicolumn{6}{c}{\textbf{Small Scale Models (\textless 7B)}} \\
\cmidrule(lr){3-5} \cmidrule(lr){6-11}
& \textbf{Size} & \textbf{Long-VITA-128K} & \textbf{LongLLAVA} & \textbf{Qwen2.5-VL} & \textbf{InternVL2.5} & \textbf{Ovis2} & \textbf{GlimpsePrune} & \textbf{Qwen2.5-VL} & \textbf{VisionZip} & \textbf{VLZip*} \\
& (\textbf{Params}) & \textbf{14B} & \textbf{9B} & \textbf{7B} & \textbf{4B} & \textbf{4B} & \textbf{3B} & \textbf{3B} & \textbf{3B} & \textbf{3B} \\
\midrule
\multirow{5}{*}{VRAG} 
& 8k   & 50.3 & 31.9 & 48.3 & 40.9 & 35.0 & 40.6 & \textbf{43.4} & \underline{41.7} & 25.2 \\
& 16k  & 50.9 & 31.3 & 45.3 & \underline{38.6} & 30.7 & 37.4 & 38.1 & \textbf{38.7} & 28.1 \\
& 32k  & 41.8 & 28.1 & 43.9 & 30.5 & 29.4 & 33.2 & \underline{34.4} & \textbf{35.3} & 26.3 \\
& 64k  & 46.5 & 27.1 & 35.1 & 30.9 & 31.7 & 28.7 & \textbf{34.6} & \underline{33.6} & 27.0 \\
& 128k & -    & 18.9 & 31.1 & \underline{21.3} & 4.7  & 8.2  & 10.8 & 8.6  & \textbf{23.3} \\
\midrule
\multirow{5}{*}{NIAH} 
& 8k   & 63.1 & 48.7 & 56.1 & \textbf{59.0} & 49.0 & \underline{56.2} & 55.3 & 52.3 & 35.7 \\
& 16k  & -    & 47.6 & 52.1 & \textbf{55.2} & 37.9 & \underline{51.7} & 50.5 & -    & 32.1 \\
& 32k  & -    & 42.4 & 45.6 & \textbf{49.2} & 29.5 & \underline{44.2} & 43.3 & -    & 30.8 \\
& 64k  & -    & 39.5 & 33.8 & \textbf{41.8} & 26.9 & 34.3 & \underline{35.2} & -    & 28.5 \\
& 128k & -    & 36.5 & 27.0 & \underline{25.4} & 13.2 & 13.5 & 13.4 & -    & \textbf{25.3} \\
\midrule
\multirow{5}{*}{ICL}  
& 8k   & -    & 76.5 & 97.5 & \underline{93.1} & 64.8 & 90.6 & \underline{93.1} & 91.2 & \textbf{94.1} \\
& 16k  & -    & 52.4 & 91.2 & \underline{78.6} & 13.8 & 55.3 & 68.1 & -    & \textbf{87.5} \\
& 32k  & -    & 39.3 & 81.5 & \underline{51.0} & 5.5  & 16.5 & 20.0 & -    & \textbf{79.8} \\
& 64k  & -    & 16.8 & 57.5 & \underline{8.8}  & 0.5  & 4.8  & 7.3  & -    & \textbf{63.0} \\
& 128k & -    & -    & 44.0 & 0.8  & 0.5  & 6.8  & \underline{7.5}  & -    & \textbf{58.8} \\
\bottomrule
\end{tabular}%
}
\end{table*}

\noindent\textbf{Results on MMLongBench.}
To further validate generalization, we evaluate on three MMLongBench tasks: visual retrieval-augmented generation (VRAG), needle-in-a-haystack (NIAH), and many-shot in-context learning (ICL). While showing a modest performance trade-off on shorter-context retrieval tasks compared to the baseline—a characteristic of its specialization, which we analyze in the sequel—Tab.~\ref{tab:model_performance_mmlongbench} demonstrates that VLZip's advantages become increasingly pronounced at extreme context lengths.
At 128k tokens, where most models degrade severely, VLZip achieves \textbf{23.3} on VRAG and \textbf{25.3} on NIAH, ranking first among all models of comparable scale ($<$7B parameters) and greatly outperforming other compression methods like GlimpsePrune and VisionZip.
Most strikingly, VLZip achieves \textbf{58.8} on ICL-128k, substantially outperforming all competing models including Qwen2.5-VL-7B (44.0)---a model twice its size.
This exceptional ICL performance directly validates our multi-layer injection design: by continuously providing fine-grained global context to every decoder layer, VLZip retains the ability to leverage long-range in-context examples even at extreme lengths where other methods collapse.

\begin{figure}[t] 
    \centering 
    \begin{subfigure}{0.48\linewidth}
        \includegraphics[width=\linewidth]{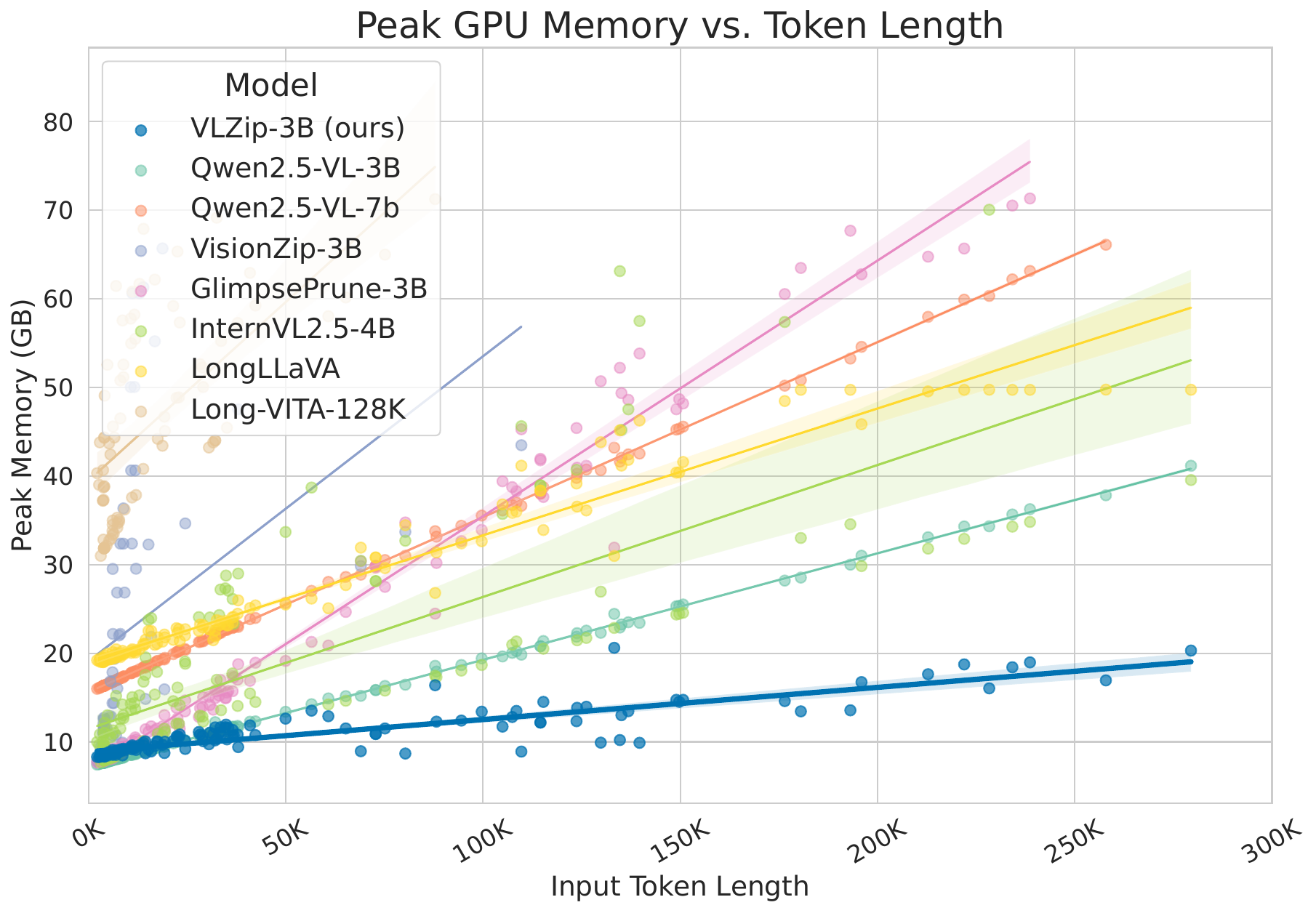}
        \caption{The peak GPU memory usage versus input token length on the LongVLBench dataset.}
        \label{fig:memory_vs_token} 
    \end{subfigure}
    \hfill 
    \begin{subfigure}{0.48\linewidth}
        \includegraphics[width=\linewidth]{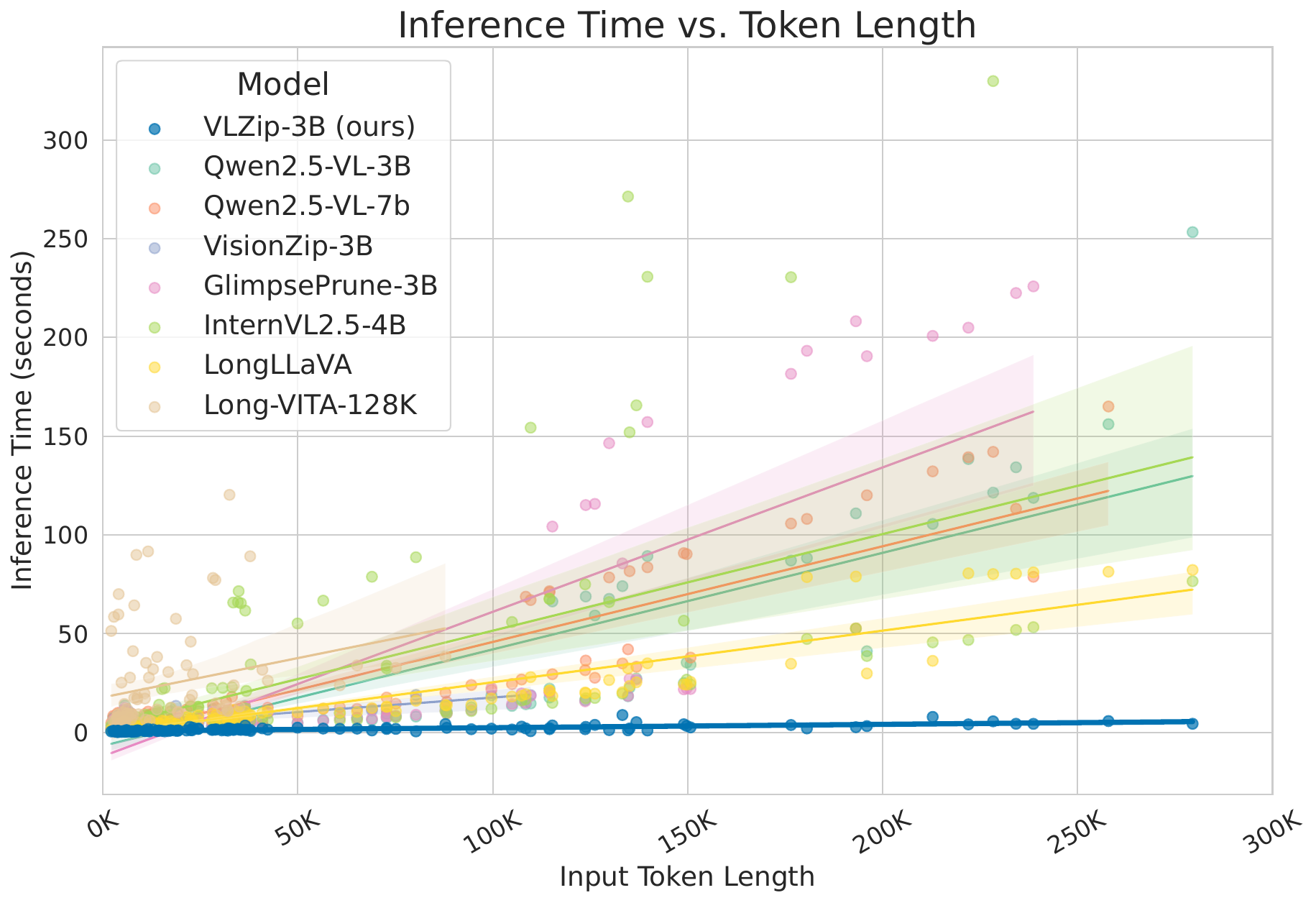}
        \caption{The inference time versus input token length on the LongVLBench dataset.}
        \label{fig:time_vs_token} 
    \end{subfigure}
    \caption{Analysis of resource consumption (memory and time) versus input token length on the LongVLBench dataset.}
    \label{fig:resource_analysis} 
\end{figure}

\begin{wrapfigure}{r}{0.46\linewidth}  
    \centering
    \includegraphics[width=0.44\textwidth]{{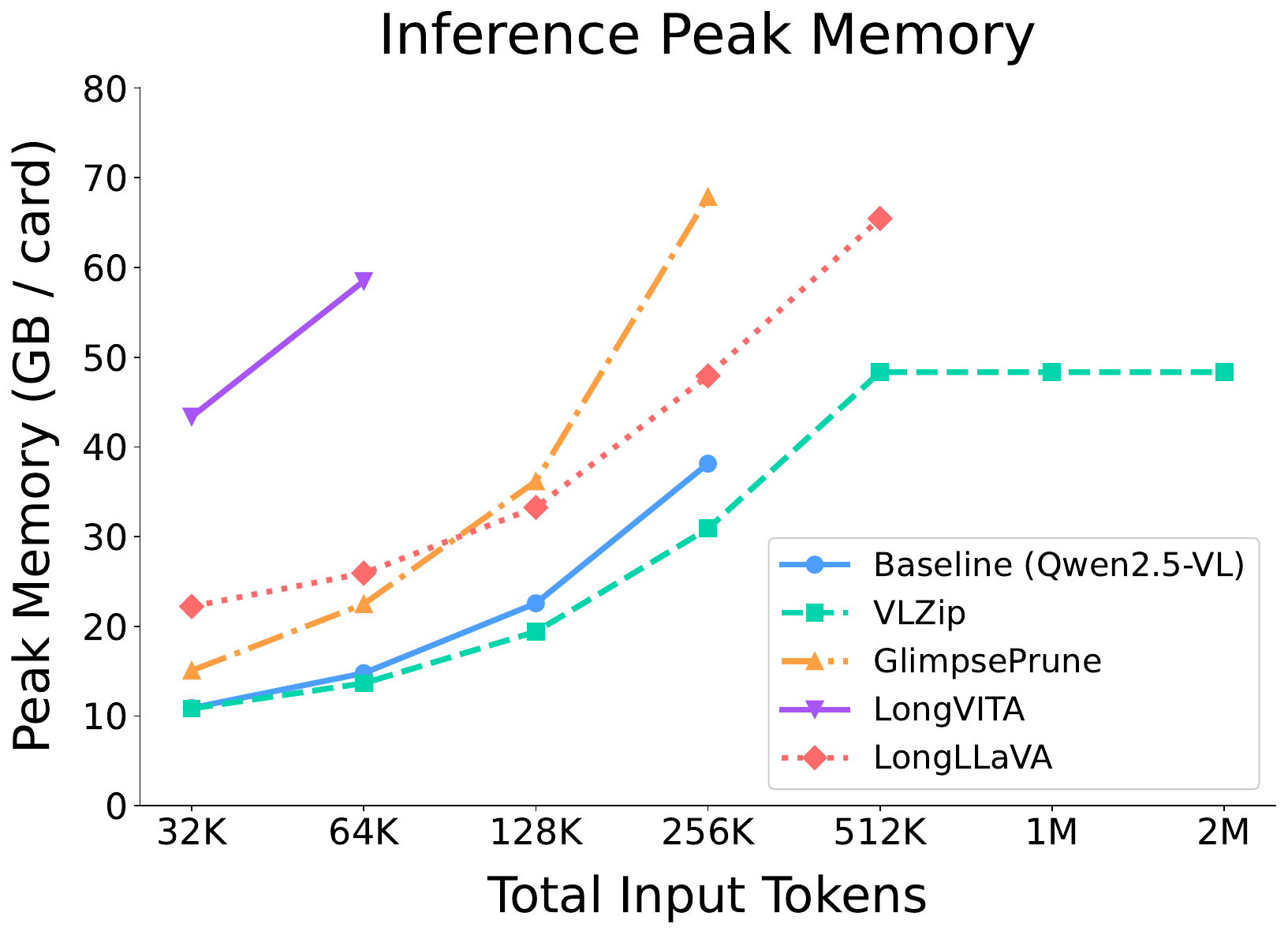}}
    \caption{Prefilling peak memory vs.\ input token length.}
    \label{fig:peak_mem}
\end{wrapfigure}

\noindent\textbf{Inference Efficiency and Scalability.}
Our architectural design translates directly into major computational gains. As shown in Figs.~\ref{fig:memory_vs_token} and~\ref{fig:time_vs_token}, peak GPU memory scales significantly more favorably, enabling the processing of sequences over 280k tokens on a single A100-80GB GPU—a scale unattainable by baselines. Inference speed similarly benefits, with VLZip running 23.6$\times$ faster than the base model at extremely long context lengths.
To further isolate memory scaling behavior, we measure prefilling peak memory on synthetic interleaved sequences with a 1:1 image-to-text token ratio across lengths from 32K to 2M tokens (Fig.~\ref{fig:peak_mem}). All baselines exhibit steep memory growth: Long-VITA exceeds the 80GB A100 limit before 128K tokens, GlimpsePrune surpasses it near 256K, and LongLLaVA reaches the limit around 512K tokens. The uncompressed baseline (Qwen2.5-VL) similarly runs out of memory beyond 256K tokens. In contrast, VLZip's memory curve flattens after 512K tokens by batch-processing images through the visual compressor before decoding, amortizing the visual encoding cost. This allows VLZip to scale to \textbf{2M} tokens—an order of magnitude beyond all baselines—while keeping peak memory under 50GB on a single card. Together, the minimal memory overhead and fast inference establish VLZip as a practical solution for long-context applications.

\noindent\textbf{Short-Context Performance.}
To assess the impact of our specialization, we evaluated VLZip on standard short-context benchmarks (Tab.~\ref{tab:multimodal_benchmark}). Since these benchmarks involve only short textual queries, text compression is not activated, and only visual compression is applied. As anticipated, and consistent with the trade-offs observed on shorter-context MMLongBench tasks, the compression-focused design introduces a moderate trade-off on perception-heavy tasks such as GQA and AI2D, while performance remains competitive on knowledge-heavy VQA tasks, with VLZip even surpassing the baseline on OKVQA. Overall, VLZip retains over 90\% of the baseline's performance across most benchmarks, confirming that long-context compression capability is achieved without fundamentally compromising core vision-language abilities.

\begin{table*}[t]
\centering
\caption{Performance on short-context VLM benchmarks: 
MMBench~\cite{liu2024mmbench}, GQA~\cite{hudson2019gqa}, 
POPE~\cite{li2023evaluating_pope}, AI2D~\cite{kembhavi2016diagram},
MMMU~\cite{yue2024mmmu}, SEED-Bench-Image (SEED$^{\text{img}}$)~\cite{li2024seed},
ScienceQA-Image (SQA$^{\text{img}}$)~\cite{lu2022learn_sqa}, 
and OKVQA~\cite{marino2019okvqa}.}

\label{tab:multimodal_benchmark}
\resizebox{0.9\textwidth}{!}{
\begin{tabular}{l | cccccccc}
\toprule
Model & MMBench & GQA & POPE & AI2D & MMMU & SEED$^{\text{img}}$ & SQA$^{\text{img}}$ & OKVQA \\
\midrule
Qwen2.5-VL-3B  & 78.4 & 60.0 & 87.6 & 78.6 & 46.9 & 74.8 & 81.1 & 42.5 \\
VLZip-3B       & 72.9 & 53.8 & 83.4 & 71.6 & 43.1 & 67.9 & 79.2 & 51.1 \\
\bottomrule
\end{tabular}
}
\end{table*}

\subsection{Ablation Studies}
Next, we conduct a series of ablation studies to dissect VLZip's core components and validate our key design choices.

\noindent\textbf{Necessity of Unified Compression.}
We validated the necessity of our compression modules by measuring the maximum trainable sequence length on 8×A100 GPUs with DeepSpeed ZeRO-2, tested in steps of 10K tokens (Table~\ref{tab:module_necessity}). The uncompressed baseline model fails beyond 20K tokens due to Out-of-Memory (OOM) errors. Compressing a single modality offers only partial relief: visual compression alone extends the limit to 50K tokens, while textual compression \begin{wraptable}{r}{0.45\linewidth}
\centering
\caption{Impact of compression modules on maximum trainable sequence length before OOM (K = 1024 tokens).}
\label{tab:module_necessity}
\small
\setlength{\tabcolsep}{6pt}
\begin{tabular}{cc|c}
\toprule
\multicolumn{2}{c|}{\textbf{Compression}} & \textbf{Max. Len.} \\
\textbf{Visual} & \textbf{Textual} & \textbf{(K tokens)} \\
\midrule
\ding{55} & \ding{55} & 20 \\
\ding{51} & \ding{55} & 50 \\
\ding{55} & \ding{51} & 40 \\
\ding{51} & \ding{51} & \textbf{120} \\
\bottomrule
\end{tabular}
\end{wraptable}reaches 40K tokens, indicating that both modalities contribute significantly to memory pressure in ultra-long multimodal sequences. In stark contrast, our unified approach expands this limit to 120K tokens---\textbf{6-fold} the baseline capacity. This demonstrates that compressing both modalities in unison is not merely beneficial but essential for enabling direct training on ultra-long contexts.

\noindent\textbf{Training Curriculum Analysis.}
Tab.~\ref{tab:training_stages_effect} reveals the distinct role of each stage. Comparing Rows 1 and 4, adding Stage 4 (long-context consolidation) substantially improves ICL-128k and LongVLBench average, confirming its importance for ultra-long reasoning. Row 2 shows that skipping Stage 3 (joint interleaved fine-tuning) causes ICL to collapse, indicating joint fine-tuning is essential for harmonizing both modalities. Row 3 demonstrates that even without compressor pretraining (Stages 1\&2), the model achieves competitive LongVLBench scores, but the full pipeline (Row 4) yields the best overall balance across all tasks.

\begin{table}[t]
\centering
\caption{Ablation study on training stages. We report both average and 128k scores on four tasks to show each stage's contribution. The full pipeline is marked with \colorbox{gray!20}{gray}. \textbf{Bold} indicates the best.}
\label{tab:training_stages_effect}
\setlength{\tabcolsep}{4pt}
\small
\begin{tabular}{cccc|cc|cc|cc|cc}
\toprule
\multicolumn{4}{c|}{Training Stage} 
  & \multicolumn{2}{c|}{VRAG} 
  & \multicolumn{2}{c|}{NIAH} 
  & \multicolumn{2}{c|}{ICL} 
  & \multicolumn{2}{c}{LongVLBench} \\
S1 & S2 & S3 & S4 & avg & 128k & avg & 128k & avg & 128k & avg & $>$128k \\
\midrule
\ding{51} & \ding{51} & \ding{51} & \ding{55} & 22.7 & 22.8 & \textbf{30.9} & \textbf{28.1} & 71.8 & 46.8 & 59.4 & 7.05 \\
\ding{51} & \ding{51} & \ding{55} & \ding{51} & \textbf{26.4} & 22.8 & 25.8 & 22.9 & 24.7 & 4.3 & 59.1 & 6.25 \\
\ding{55} & \ding{55} & \ding{51} & \ding{51} & 23.6 & 22.2 & 29.4 & 24.7 & 73.1 & 47.8 & 58.6 & \textbf{7.60} \\
\rowcolor{gray!15}
\ding{51} & \ding{51} & \ding{51} & \ding{51} & 26.0 & \textbf{23.3} & 30.5 & 25.3 & \textbf{76.6} & \textbf{58.8} & \textbf{61.9} & \textbf{7.60} \\
\bottomrule
\end{tabular}
\end{table}

\begin{table}[t]
\centering
\caption{Ablation on chunk size and tokens per chunk. 
We set $C_v = C_t = C$ and $M_v = M_t = M$ across all configurations, 
and fix the full-layer injection strategy. 
The default configuration ($C{=}100$, $M{=}4$) is marked with 
\colorbox{gray!20}{gray}. \textbf{Bold} indicates the best.}
\label{tab:ablation_cs_m}
\setlength{\tabcolsep}{4pt}
\small
\begin{tabular}{cc|cc|cc|cc|cc}
\toprule
\multirow{2}{*}{$C$} & \multirow{2}{*}{$M$} 
  & \multicolumn{2}{c|}{VRAG} 
  & \multicolumn{2}{c|}{NIAH} 
  & \multicolumn{2}{c|}{ICL} 
  & \multicolumn{2}{c}{LongVLBench} \\
& & avg & 128k & avg & 128k & avg & 128k & avg & $>$128k \\
\midrule
\multicolumn{10}{l}{\textit{Varying tokens per chunk $M$ (fixed $C{=}100$)}} \\
\midrule
100 & 2 & 28.7 & \textbf{29.0} & 27.8 & 25.0 & 69.1 & 39.3 & 58.8 & 7.45 \\
\rowcolor{gray!15}
100 & 4 & 26.0 & 23.3 & 30.5 & \textbf{25.3} & \textbf{76.6} & \textbf{58.8} & 61.9 & \textbf{7.60} \\
100 & 8 & 26.7 & 23.1 & 31.1 & 25.0 & 64.1 & 20.5 & 60.3 & 7.15 \\
\midrule
\multicolumn{10}{l}{\textit{Varying chunk size $C$ (fixed $M{=}4$)}} \\
\midrule
25  & 4 & \textbf{30.7} & 28.0 & 31.4 & 24.5 & 58.8 & 12.8 & \textbf{63.3} & 7.05 \\
50  & 4 & 26.6 & 25.1 & \textbf{32.8} & 24.0 & 59.7 & 22.8 & 59.4 & 6.05 \\
\rowcolor{gray!15}
100 & 4 & 26.0 & 23.3 & 30.5 & \textbf{25.3} & \textbf{76.6} & \textbf{58.8} & 61.9 & \textbf{7.60} \\
\midrule
\multicolumn{10}{l}{\textit{Joint variation}} \\
\midrule
50  & 8 & 29.7 & 27.2 & 31.6 & 23.4 & 49.0 &  7.5 & 62.1 & 7.00 \\
\bottomrule
\end{tabular}
\end{table}

\noindent\textbf{Ablation on Chunk Size and Tokens per Chunk.}
To reduce the hyperparameter search space, we tie the image and text parameters throughout this ablation, setting $C_v {=} C_t {=} C$ and $M_v {=} M_t {=} M$, leaving asymmetric configurations as future work. We ablate $C \in \{25, 50, 100\}$ and $M \in \{2, 4, 8\}$ under the full-layer injection strategy. 
Results are shown in Tab.~\ref{tab:ablation_cs_m}. Fixing $M{=}4$ and varying $C$, we observe that smaller chunk sizes ($C=25$) tend to improve performance on retrieval-centric tasks (VRAG and NIAH),
as finer-grained compression preserves more local retrieval cues within each chunk. However, this advantage reverses at ultra-long contexts: $C{=}100$ achieves the best LongVLBench $>$128k score (7.60) and by far the highest ICL-128k score (58.8). We attribute this to the fact that larger chunks capture broader contextual patterns within each compressed unit, which is critical for tasks requiring global sequence understanding at extreme lengths.
Fixing $C{=}100$ and varying $M$, we find $M{=}4$ to be the optimal choice across all ultra-long metrics. Increasing to $M{=}8$ does not further improve performance and degrades ICL-128k significantly (20.5 vs.\ 58.8), while decreasing to $M{=}2$ reduces NIAH average (27.8 vs.\ 30.5) and ICL-128k (39.3 vs.\ 58.8), suggesting insufficient capacity to represent compressed chunk information. We also examine a joint variation ($C{=}50$, $M{=}8$), which maintains a higher compression rate than our default, but yields noticeably worse ICL performance (ICL avg: 49.0, ICL-128k: 7.5), further confirming that $C$ and $M$ have independent effects that cannot be captured by compression rate alone. We therefore adopt $C{=}100$, $M{=}4$ as our default configuration.

\noindent\textbf{Injection Layer Strategy.}
Tab.~\ref{tab:ablation_injection} compares four injection strategies under our default $C{=}100$, $M{=}4$ setting. First-layer injection yields the strongest retrieval performance (VRAG avg: 31.2, NIAH avg: 31.3), as concentrating compressed features at the earliest layer provides rich context at the input stage.\begin{table}[t]
\centering
\caption{Injection layer strategy ablation for $C{=}100$, $M{=}4$.
\textbf{Bold} denotes the best.}
\label{tab:ablation_injection}
\setlength{\tabcolsep}{4pt}
\small
\begin{tabular}{l|cc|cc|cc|cc}
\toprule
\multirow{2}{*}{Strategy} 
  & \multicolumn{2}{c|}{VRAG} 
  & \multicolumn{2}{c|}{NIAH} 
  & \multicolumn{2}{c|}{ICL}
  & \multicolumn{2}{c}{LongVLBench} \\
& avg & 128k & avg & 128k & avg & 128k & avg & $>$128k \\
\midrule
First-layer  & \textbf{31.2} & \textbf{31.3} & \textbf{31.3} & \textbf{27.8} & 75.8 & 55.5 & 59.6 & 8.00 \\
First-half   & 30.0 & 28.0 & 28.9 & 27.1 & 70.2 & 42.3 & 60.6 & 7.65 \\
Interval-2   & 28.9 & 27.8 & 30.6 & 27.2 & 70.8 & 39.3 & 61.8 & \textbf{8.25} \\
\rowcolor{gray!15}
Full-layer   & 26.0 & 23.3 & 30.5 & 25.3 & \textbf{76.6} & \textbf{58.8} & \textbf{61.9} & 7.60 \\
\bottomrule
\end{tabular}
\end{table}
However, full-layer injection achieves the best ICL scores (ICL avg: 76.6, ICL-128k: 58.8) and LongVLBench average (61.9). While other strategies show competitive point scores in the longest bin, the superior average and ICL performance of full-layer injection confirms its robustness. We attribute this to the holistic nature of these tasks: ICL requires inferring patterns from interleaved multimodal examples distributed across the entire context, while LongVLBench emphasizes comprehensive reasoning over long sequences. Both demand global context understanding that benefits from compressed representations being continuously available at every decoder layer, rather than concentrated at early layers alone.

\begin{table}[t]
\centering
\caption{Ablation on compressor architecture (fixed C=100, M=4, full-layer injection). The Q-Former compressor is marked with \colorbox{gray!20}{gray}. \textbf{Bold} indicates the best.}

\label{tab:compressor_arch}
\setlength{\tabcolsep}{4pt}
\small
\begin{tabular}{l|cc|cc|cc|cc}
\toprule
\multirow{2}{*}{Compressor} & \multicolumn{2}{c|}{VRAG} & \multicolumn{2}{c|}{NIAH} & \multicolumn{2}{c|}{ICL} & \multicolumn{2}{c}{LongVLBench} \\
 & avg & 128k & avg & 128k & avg & 128k & avg & >128k \\
\midrule
Avg Pooling & \textbf{32.4} & \textbf{32.1} & 26.3 & 25.0 & 35.8 & 6.0 & 53.8 & 6.85 \\
\rowcolor{gray!15}
Q-Former & 26.0 & 23.3 & \textbf{30.5} & \textbf{25.3} & \textbf{76.6} & \textbf{58.8} & \textbf{61.9} & \textbf{7.60} \\
\bottomrule
\end{tabular}
\end{table}
\noindent\textbf{Compressor Architecture.}
Tab.~\ref{tab:compressor_arch} compares our Q-Former compressor with a simpler average pooling baseline. Average pooling achieves stronger VRAG retrieval scores, as averaging preserves distributional cues beneficial for matching-based tasks. However, it severely underperforms on ICL and LongVLBench, indicating that average pooling lacks the capacity to selectively preserve structured information essential for reasoning over long contexts.

\section{Conclusion}

We introduced VLZip, a unified compression framework that enables efficient, high-fidelity reasoning over ultra-long multimodal sequences within a pure Transformer architecture. By hierarchically distilling visual and textual information and injecting it across all decoder layers, VLZip enables training on sequences up to 120K tokens—a 6$\times$ improvement—and processes contexts exceeding 280K tokens during inference. Its memory efficiency further enables scaling to 2M-token contexts, a scale infeasible for most comparable methods.

\section*{Acknowledgements}
This work is supported by Ant Group Research Fund and the National Natural Science Foundation of China (Grant No. 62472098), the Science and Technology Commission of Shanghai Municipality (No. 25511106100).

%
%
\bibliographystyle{splncs04}
\bibliography{main}
\clearpage

\appendix

\input{suppl-camera-ready}
\end{document}

%% file: suppl-camera-ready.tex
\section{Extended Related Work}
The advancement of long-context VLMs is an underexplored field, facing dual challenges in both model architecture and robust evaluation. 
In this section, we review the handful of existing works that address these challenges.

\noindent\textbf{Architectures.} The nascent field of long-context VLM architectures has primarily explored three distinct strategies, each forcing a difficult compromise between fidelity, efficiency, and performance. One path is Input Pruning, where methods like GlimpsePrune~\cite{DBLP:journals/corr/abs-2508-01548} and VisionZip~\cite{DBLP:conf/cvpr/YangCTWL0J25} permanently discard visual tokens. This "prune-and-forget" tactic risks irreversible information loss and suffers from a myopic focus on the visual modality, ignoring the equally vast textual information in interleaved sequences. A second path, Architectural Overhaul, replaces the Transformer with more efficient backbones like SSMs, as seen in LongLLaVA~\cite{DBLP:journals/corr/abs-2409-02889}. While computationally appealing, this can compromise the precise, non-local reasoning at which pure Transformers excel. Finally, Data-Centric Enhancements like Mantis~\cite{DBLP:journals/tmlr/JiangHZWKLC24} improve multi-image reasoning but do so by trading efficiency for performance, failing to address the fundamental scaling problem for ultra-long contexts.

A promising alternative to these hard compromises has emerged from the NLP community in the form of Soft Prompt Compression. As surveyed in~\cite{li2025prompt}, methods like GIST~\cite{mu2023learning} and the In-Context Autoencoder~\cite{ge2024context} have shown that long textual contexts can be effectively distilled into a small set of compact, information-rich "soft prompts." Frameworks like xRAG~\cite{cheng2024xrag}, E2LLM~\cite{liao2025e2llm}, and AutoCompressor~\cite{chevalier2023adapting} further refine this by using a dedicated text encoder (or decoder) to generate these soft prompts. These techniques demonstrate a powerful principle: long, complex sequences can be represented by dense vectors, offering a path to efficiency without aggressive pruning. However, their utility is limited as they are designed exclusively for the text modality and typically employ a simple, single-layer injection of the compressed prompt.

VLZip builds on this principle, being the first framework to generalize soft prompt compression to the multimodal domain. It unifies the hierarchical compression of both images and text and introduces a novel multi-layer prefix injection mechanism for richer contextual conditioning throughout the VLM's decoder layers.


\noindent\textbf{Evaluations.} 
Paralleling the challenges in modeling, the evaluation of long-context VLMs is hindered by significant gaps in the few existing benchmarks. Recent efforts, such as Mantis-Eval~\cite{DBLP:journals/tmlr/JiangHZWKLC24} and MileBench~\cite{DBLP:journals/corr/abs-2404-18532}, often repurpose short-context datasets, resulting in imbalanced image-to-text ratios and total token lengths insufficient to truly test modern models. Furthermore, benchmarks like MM-NIAH~\cite{DBLP:conf/nips/WangZRDLLH0ZLZL24} and MMLongCite~\cite{DBLP:journals/corr/abs-2510-13276} rely on artificial `needle-in-a-haystack' (NIAH) retrieval tests. While effective for measuring recall, these tasks do not reflect practical scenarios that demand holistic reasoning. Another common issue is the use of filler content, as seen in MMLongBench~\cite{wang2025mmlongbenchbenchmarkinglongcontextvisionlanguage}, where long contexts are constructed by padding with external documents. This approach introduces the risk of data contamination from pre-training corpora and uses material that is not integral to the core reasoning task.

These limitations highlight the need for a benchmark constructed from the ground up to provide a more authentic and rigorous testbed. This motivates our creation of LongVLBench, which features genuinely long, coherent sequences with a high and balanced density of multimodal information, moving beyond simple retrieval to test authentic narrative understanding.


\section{Experimental Setup}

\subsection{Training Datasets}

Our training pipeline utilizes carefully curated datasets at each stage to progressively develop the model's compression and reasoning capabilities (Table~\ref{tab:training_datasets}). 

\begin{wraptable}{r}{0.5\linewidth}
\centering
\caption{Training dataset composition for each stage.}
\label{tab:training_datasets}
\resizebox{\linewidth}{!}{%
\begin{tabular}{llrr}
\toprule
\textbf{Stage} & \textbf{Dataset} & {\textbf{\# Train}} & {\textbf{\# Eval}} \\
\midrule
Stage 1 & LLaVA-OV-SI & 2549886 & 25757 \\
\midrule
Stage 2 & ChatQA2(Reconstruction) & 4928756 & 49786 \\
\midrule
\multirow{2}{*}{Stage 3} & LLaVA-OV-OV & \multicolumn{1}{r}{\multirow{2}{*}{2208445}} & \multicolumn{1}{r}{\multirow{2}{*}{22366}} \\
            & VEGA-4k & & \\ 
\midrule 
Stage 4 & ChatQA2 & 102965 & 1041 \\
\bottomrule
\end{tabular}
}

\end{wraptable}

\noindent\textbf{Stage 1: Visual Pre-training.} We employ the LLaVA-OneVision~\cite{DBLP:journals/tmlr/0080ZGZ00ZZL0L25_llavaov} Single-Image (LLaVA-OV-SI) dataset for training the hierarchical visual compressor. Pure text samples are excluded since the LLM decoder is frozen.

\noindent\textbf{Stage 2: Textual Pre-training.} For the textual compression module, we construct a self-supervised reconstruction task from the ChatQA2~\cite{xu2024chatqa} long-document corpus. Each document is first de-duplicated by input content, then segmented into fixed-length chunks of 100 tokens. Every chunk is paired with a reconstruction instruction sampled from a diverse set of prompt templates, training the compressor to preserve the semantic content of each segment.

\noindent\textbf{Stage 3: Joint Fine-tuning.} We use the LLaVA-OneVision OneVision (LLaVA-OV-OV) dataset, which provides both single-image and multi-image instruction data (excluding video samples), combined with the VEGA-4k~\cite{DBLP:journals/corr/abs-2406-10228_vega} dataset to facilitate unified compression and reasoning.

\noindent\textbf{Stage 4: Long-Context Consolidation.} We use the original ChatQA2 dataset to adapt the model's reasoning strategies to ultra-long contexts. We opt for text-only data at this stage due to the absence of ultra-long interleaved image-text training datasets in the community, while still ensuring effective leverage of compression capabilities for extended reasoning tasks.


\subsection{Training Hyperparameters}

We provide complete training hyperparameters for all four stages in Table~\ref{tab:hyperparams}. All training uses mixed precision (BF16), Flash Attention~\cite{DBLP:conf/iclr/Dao24_flashattention}, and gradient checkpointing for memory efficiency. All stages employ early stopping based on validation loss evaluated every 1000 steps. The visual and textual Q-Formers utilize standard Transformer architectures with 8 attention heads. Note that the Q-Former modules use a higher learning rate than the base rate in Stages~1 and~2 to accelerate convergence of the newly initialized parameters. The total training time across all stages is approximately 2 days (49 hours) on 32 GPUs.

\begin{table}[t]
\centering
\caption{Training hyperparameters for each stage.}
\label{tab:hyperparams}
\small
\resizebox{\textwidth}{!}{
\begin{tabular}{lcccc}
\toprule
\textbf{Hyperparameter} & \textbf{Stage 1} & \textbf{Stage 2} & \textbf{Stage 3} & \textbf{Stage 4} \\
 & Visual Pre-train & Text Pre-train & Joint Fine-tune & Long-Context \\
\midrule
\multicolumn{5}{l}{\textit{Trainable Parameters}} \\
\quad Vision Encoder & & & & \\
\quad Vision Projection Adapter & \ding{51} & & \ding{51} & \ding{51} \\
\quad Visual Q-Former & \ding{51} & & \ding{51} & \ding{51} \\
\quad Text Encoder & & \ding{51} & \ding{51} & \ding{51} \\
\quad Text Projection ($\mathbf{W}_{\text{proj}}$) & & \ding{51} & \ding{51} & \ding{51} \\
\quad Textual Q-Former & & \ding{51} & \ding{51} & \ding{51} \\
\quad LLM Decoder & & & \ding{51} & \ding{51} \\
\midrule
\multicolumn{5}{l}{\textit{Training Configuration}} \\
\quad Base Learning Rate & 1e-5 & 5e-5 & 1e-5 & 1e-5 \\
\quad Q-Former Learning Rate & 1e-4 & 5e-4 & 1e-5 & 1e-5 \\
\quad Learning Rate Schedule & \multicolumn{4}{c}{Cosine} \\
\quad Warmup Ratio & \multicolumn{4}{c}{0.03} \\
\quad Batch Size & 64 & 512 & 128 & 128 \\
\quad Training Epochs & 1 & 1 & 1 & 1 \\
\quad Early Stopping & \ding{51} & \ding{51} & \ding{51} & \ding{51} \\
\quad Optimizer & \multicolumn{4}{c}{AdamW} \\
\midrule
\multicolumn{5}{l}{\textit{Hardware \& Efficiency}} \\
\quad Number of GPUs & 32 & 32 & 32 & 32 \\
\quad GPU Type & A100-80GB & A100-80GB & A100-80GB & H20-96GB \\
\quad Training Time & $\sim$8 hours & $\sim$5 hours & $\sim$27 hours & $\sim$9 hours \\
\bottomrule
\end{tabular}
}
\end{table}

\subsection{Baseline Model Descriptions}

We compare our approach against models with distinct architectural strategies: interleaved reasoning models (Mantis~\cite{DBLP:journals/tmlr/JiangHZWKLC24}, InternVL2.5~\cite{chen2024expanding_internvl2_5}, Ovis2~\cite{lu2024ovis}), dedicated long-context models (LongLLaVA~\cite{DBLP:journals/corr/abs-2409-02889}, Long-VITA~\cite{shen2025longvita}), and other compression methods (GlimpsePrune~\cite{DBLP:journals/corr/abs-2508-01548}, VisionZip~\cite{DBLP:conf/cvpr/YangCTWL0J25}).

\noindent\textbf{LongLLaVA} employs a hybrid architecture combining Mamba and Transformers, utilizing a progressive training strategy that enables processing thousands of images on a single A100 80GB GPU.

\noindent\textbf{Long-VITA} implements context-parallel distributed inference and logits-masked language modeling to handle ultra-long multimodal inputs, with a four-stage multimodal training paradigm. Since we evaluate long-context multimodal performance, we use the Long-VITA-128K variant for comparison.

\noindent\textbf{VisionZip} is a training-free visual token pruning method that selects highly informative visual tokens based on attention scores and drops the remaining. For fair comparison, we use the same backbone in our main experiments.

\noindent\textbf{GlimpsePrune} is a dynamic visual token pruning framework that introduces a glimpse token during the prefilling stage which helps to identify the irrelevant tokens. After a selected layer, these irrelevant tokens are removed.

\section{Extended Results and Ablations}

\subsection{Scaling to Larger Models}

To validate the generality of the VLZip framework beyond a single model scale, we apply it to the larger Qwen2.5-VL-7B-Instruct backbone. Crucially, this experiment is designed as a \textbf{zero-effort transfer}: we reuse the \emph{identical} training pipeline, hyperparameters, and datasets from the 3B configuration without any 7B-specific tuning. This deliberately stringent setup isolates the contribution of VLZip's \emph{architecture} from scale-specific optimization, providing a lower-bound estimate of its effectiveness at larger scales.

\begin{table}[t]
\centering
\caption{Scaling VLZip to a 7B backbone with an identical training pipeline. All results use the same hyperparameters and data as the 3B configuration.}
\label{tab:scaling_7b}
\begin{tabular}{l|cc|cc|cc|cc}
\toprule
\multirow{2}{*}{Model} & \multicolumn{2}{c|}{VRAG} & \multicolumn{2}{c|}{NIAH} & \multicolumn{2}{c|}{ICL} & \multicolumn{2}{c}{LongVLBench} \\
& avg & 128k & avg & 128k & avg & 128k & avg & {>128k} \\
\midrule
Qwen2.5-VL-7B & 40.7 & 31.1 & 42.9 & 27.0 & 74.3 & 44.0 & 47.1 & 1.6 \\
VLZip-7B       & 33.0 & 31.9 & 29.1 & 25.2 & 67.2 & 32.5 & 62.9 & 7.05 \\
\bottomrule
\end{tabular}%
\end{table}

As shown in Table~\ref{tab:scaling_7b}, VLZip's core advantage---robust reasoning at ultra-long contexts---transfers consistently to the 7B scale. On LongVLBench, VLZip-7B achieves 62.9 (\textbf{+33.5\%} over the baseline's 47.1), with the gap widening at extreme lengths: VLZip-7B scores 7.05 on inputs exceeding 128k tokens while the baseline collapses to 1.6. On VRAG-128k, VLZip-7B also outperforms the baseline (31.9 vs.\ 31.1). These patterns are consistent with the 3B results, confirming that the unified compression architecture is the primary driver of long-context capability.

On shorter-context tasks, VLZip-7B shows a moderate trade-off on VRAG and NIAH averages, and ICL-128k (32.5) is lower than the baseline (44.0). We attribute these gaps to the suboptimal training configuration: all hyperparameters were tuned for the 3B backbone and transferred without adjustment. The 7B backbone likely requires scale-specific learning rate schedules or longer Stage~4 training for the compression modules to fully align with the decoder. Despite this, VLZip-7B achieves the strongest LongVLBench score among all evaluated models, demonstrating that our framework generalizes across model scales without architectural modification. Scale-specific optimization is left as future work.

\subsection{Component Isolation}
\label{sec:component_isolation}

To verify that VLZip's gains stem from \emph{unified} compression rather than either modality alone, we isolate each compression module's contribution on MMLongBench at 128k. We compare the uncompressed Qwen2.5-VL-3B baseline against three compressed variants: \emph{Image-only}, which compresses only visual segments; \emph{Text-only}, which compresses only text segments; and the full \emph{VLZip}, which compresses both. Each variant is trained under the applicable subset of our four-stage curriculum: Image-only uses Stages 1 and 3, omitting Stage 4 because its ultra-long text-only data exceeds memory once text compression is disabled; Text-only uses Stages 2, 3, and 4; and VLZip uses the full pipeline. The results are reported in Table~\ref{tab:ablate_modality}.

\begin{table}[t]
\centering
\caption{Single-modality versus unified compression on MMLongBench at 128k.}
\small
\renewcommand{\arraystretch}{0.9}
\setlength{\tabcolsep}{6pt}
\begin{tabular}{lccc}
\toprule
Method & VRAG$_{128k}$ & NIAH$_{128k}$ & ICL$_{128k}$ \\
\midrule
Qwen2.5-VL-3B & 10.8 & 13.4 & 7.5  \\
Image-only    & 1.4  & 18.7 & 50.5 \\
Text-only     & 34.4 & 22.3 & 1.3  \\
VLZip         & 23.3 & 25.3 & 58.8 \\
\bottomrule
\end{tabular}
\label{tab:ablate_modality}
\end{table}

Compressing only one modality leads to task-specific collapse. The Image-only variant excels on ICL (50.5), where dense visual examples dominate, yet collapses on VRAG (1.4), where textual retrieval is essential. Text-only exhibits the mirror image, recovering VRAG (34.4) but collapsing on ICL (1.3). Only the unified VLZip sustains strong, balanced performance across all three tasks, confirming that the training-length and reasoning gains of our design cannot be obtained by single-modality compression alone.

\subsection{Detailed MMLongBench Scores}

\begin{figure*}[t]
\includegraphics[width=1.0\linewidth]{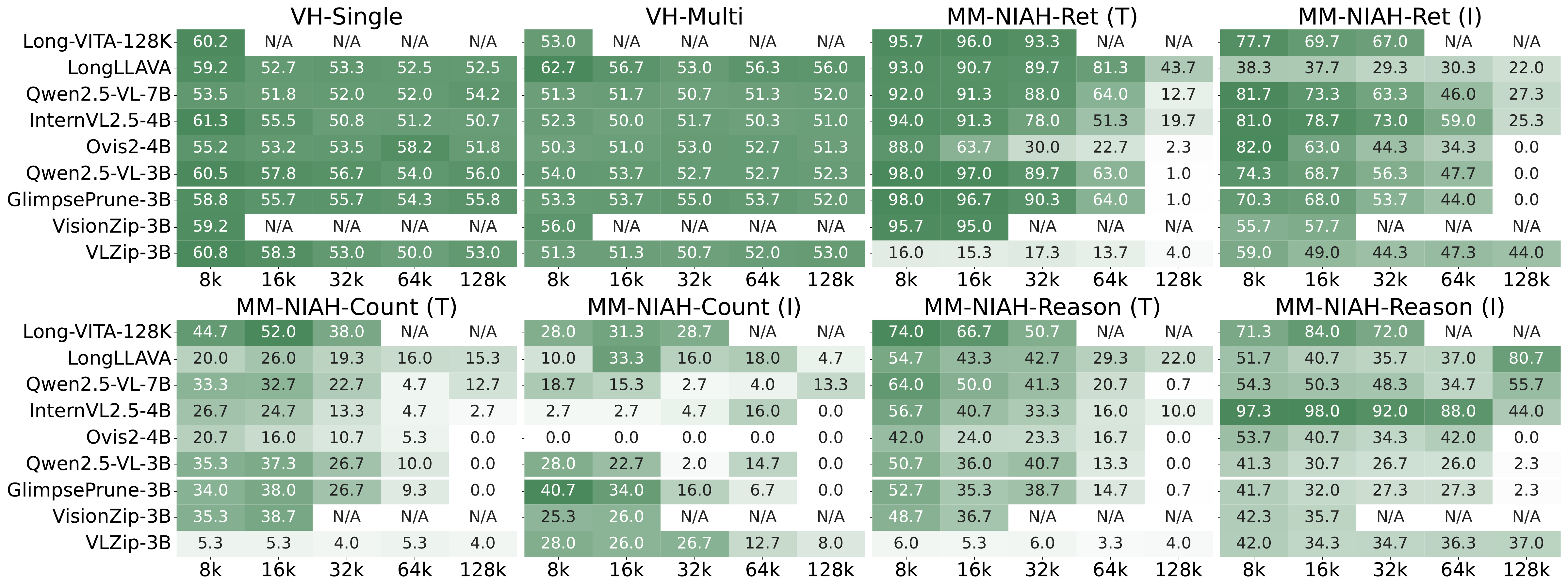}
\caption{Detailed results of models on NIAH task of MMLongBench at various lengths.}
\label{fig:NIAH}
\end{figure*}

\begin{figure*}[t]
\includegraphics[width=1.0\linewidth]{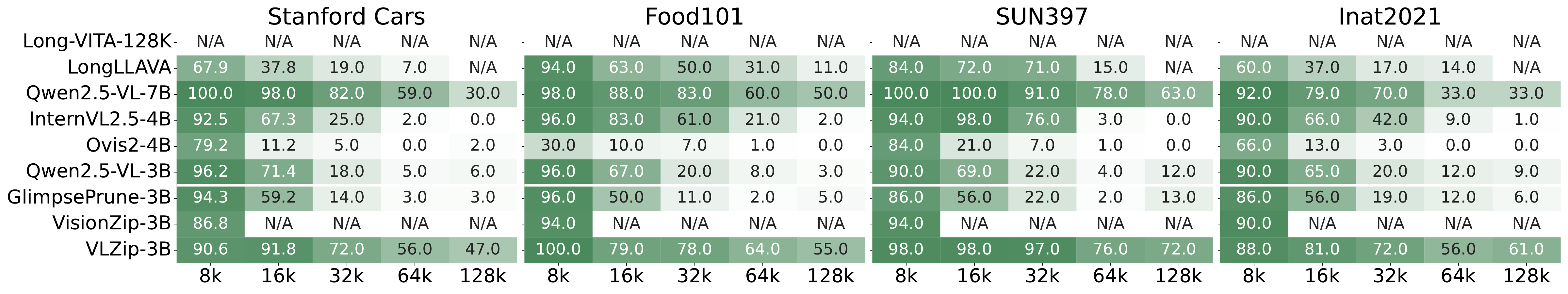}
\caption{Detailed results of models on ICL task of MMLongBench at various lengths.}
\label{fig:ICL}
\end{figure*}

\begin{wrapfigure}{r}{0.6\linewidth}  
    \centering
    \includegraphics[width=0.58\textwidth]{{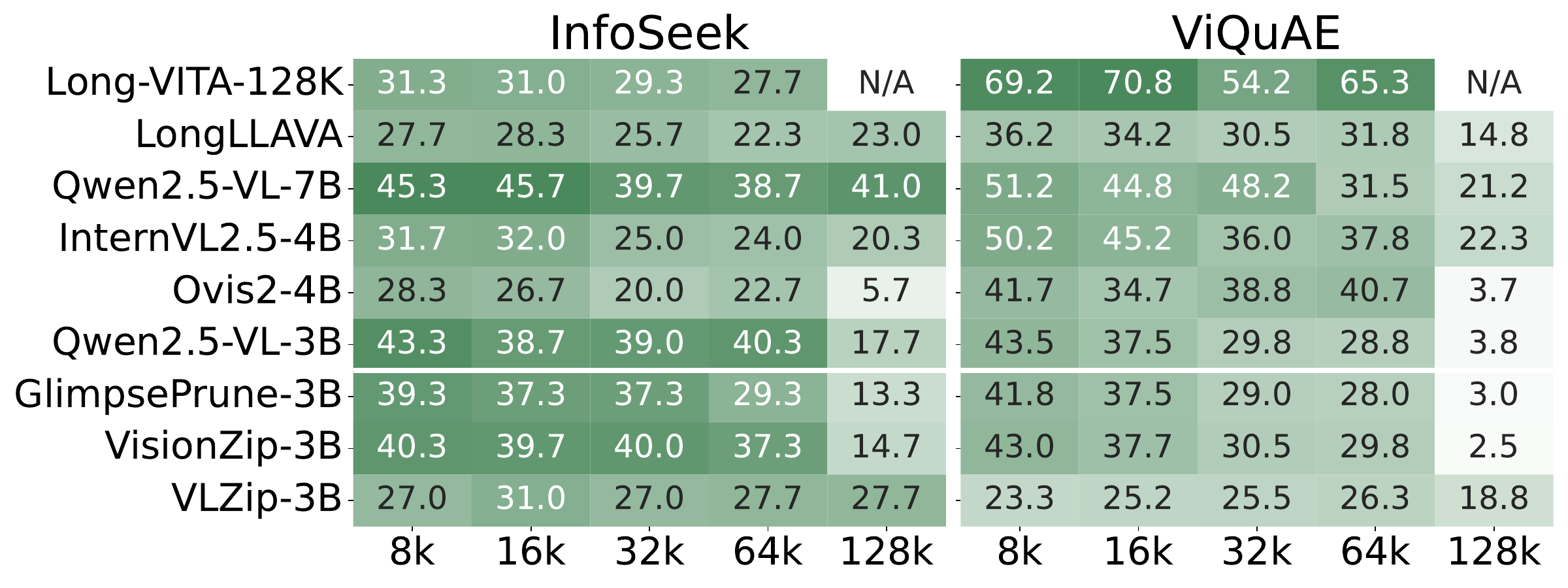}}
    \caption{Detailed results of models on VRAG task of MMLongBench at various lengths.}
    \label{fig:VRAG}
\end{wrapfigure}

We provide detailed per-task results for MMLongBench in Figures~\ref{fig:NIAH}, \ref{fig:ICL}, and \ref{fig:VRAG}, which break down the average scores reported in the main paper across different context lengths and task categories.

A notable observation from the NIAH breakdown (Figure~\ref{fig:NIAH}) is that VLZip exhibits relatively weaker performance on \emph{text-centric} NIAH subtasks (e.g., MM-NIAH-Ret~(T), MM-NIAH-Reason~(T)) compared to baselines, while remaining competitive on image-centric variants. Since NIAH tasks require locating and retrieving specific pieces of information, this pattern suggests that our default ``always compress'' strategy---which compresses all text segments unconditionally---may discard fine-grained lexical cues that are particularly important for text-based retrieval at shorter context lengths. This motivates a closer investigation into \emph{when} text compression should be activated.

\noindent\textbf{Text Compression Strategy.}
To disentangle the effect of text compression from visual compression, we compare three strategies using the same trained model: (1)~\textbf{Always}: compress all text segments by default, except for trivially short segments (under 100 characters) that are passed through uncompressed; (2)~\textbf{Never}: compress only images, leaving all text uncompressed; and (3)~\textbf{Adaptive}: apply visual compression first, then activate text compression only when the resulting sequence length exceeds the backbone's native context window (32k tokens for Qwen2.5-VL).

\begin{table}[t]
\centering
\small
\caption{Ablation on text compression strategy. "Always" compresses all text; "Never" applies image-only compression; "Adaptive" activates text compression only when the post-visual-compression sequence exceeds 32k tokens.}
\label{tab:text_compression_strategy}
\resizebox{0.95\textwidth}{!}{
\begin{tabular}{l|cccccc|cccccc|cccccc}
\toprule
\multirow{2}{*}{Strategy} & \multicolumn{6}{c|}{VRAG} & \multicolumn{6}{c|}{NIAH} & \multicolumn{6}{c}{ICL} \\
& 8k & 16k & 32k & 64k & 128k & avg & 8k & 16k & 32k & 64k & 128k & avg & 8k & 16k & 32k & 64k & 128k & avg \\
\midrule
Always   & 25.2 & 28.1 & 26.3 & 27.0 & 23.3 & 26.0 & 35.7 & 32.1 & 30.8 & 28.5 & 25.3 & 30.5 & 94.1 & 87.5 & 79.8 & 63.0 & 58.8 & 76.6 \\
Never    & 41.6 & 38.3 & 31.7 & 22.2 & 10.5 & 28.8 & 49.7 & 47.2 & 44.6 & 29.4 & 21.6 & 38.5 & 94.2 & 89.7 & 76.8 & 61.8 & 57.0 & 75.9 \\
Adaptive & 41.6 & 38.3 & 31.7 & 27.0 & 23.3 & 32.3 & 49.7 & 47.2 & 44.6 & 27.1 & 24.4 & 38.6 & 93.6 & 89.0 & 78.0 & 65.3 & 52.8 & 75.7 \\
\bottomrule
\end{tabular}
}
\end{table}

Results in Table~\ref{tab:text_compression_strategy} reveal a clear trade-off between short- and long-context performance governed by the text compression strategy. The "Never" strategy preserves full textual fidelity and achieves the strongest short-context scores (e.g., VRAG-8k: 41.6, NIAH-8k: 49.7), directly explaining the text-NIAH gap observed in Figure~\ref{fig:NIAH}. However, without text compression, performance degrades sharply at extreme lengths (VRAG-128k: 10.5, NIAH-128k: 21.6), as the uncompressed sequence exceeds the model's effective capacity. The "Always" strategy exhibits the opposite profile: it sacrifices short-context retrieval but sustains robust performance at 128k (VRAG: 23.3, NIAH: 25.3) and achieves the strongest ICL scores (avg: 76.6, 128k: 58.8), confirming that text compression is essential for ultra-long reasoning.

The "Adaptive" strategy effectively combines the advantages of both by deferring text compression until the post-visual-compression sequence exceeds 32k tokens. At short lengths ($\leq$32k), it closely matches the "Never" strategy; beyond 32k, it recovers the long-context robustness of "Always". This yields the most balanced overall results: the highest VRAG average (32.3) and NIAH average (38.6), with competitive ICL scores across all lengths. We adopt "Always" as the default in our main experiments for its simplicity and strongest ultra-long performance, while noting that the "Adaptive" variant is a practical alternative when balanced performance across all context lengths is desired.

\subsection{Failure Case Analysis on GQA}
\label{sec:failure_gqa}

To better understand the moderate trade-off VLZip exhibits on perception-heavy short-context tasks, we examine its errors on GQA in detail. 
Across the full evaluation set, there are 1{,}777 samples (14.1\% of the total) that VLZip answers incorrectly while the Qwen2.5-VL-3B baseline answers correctly.
A closer inspection reveals that a large fraction of these are not genuine reasoning failures. Specifically, 44.8\% are exact opposite-pair reversals (e.g., yes$\leftrightarrow$no, left$\leftrightarrow$right), and a further 13.1\% are semantically equivalent responses (e.g., \emph{couch} versus \emph{chair}) that are nonetheless penalized under exact-match scoring. Together these account for nearly 60\% of these baseline-correct errors, indicating that much of the measured degradation reflects fine-grained perceptual or lexical mismatch rather than a breakdown in semantic understanding.

\begin{figure}[t]
    \centering
    \includegraphics[width=1.0\linewidth]{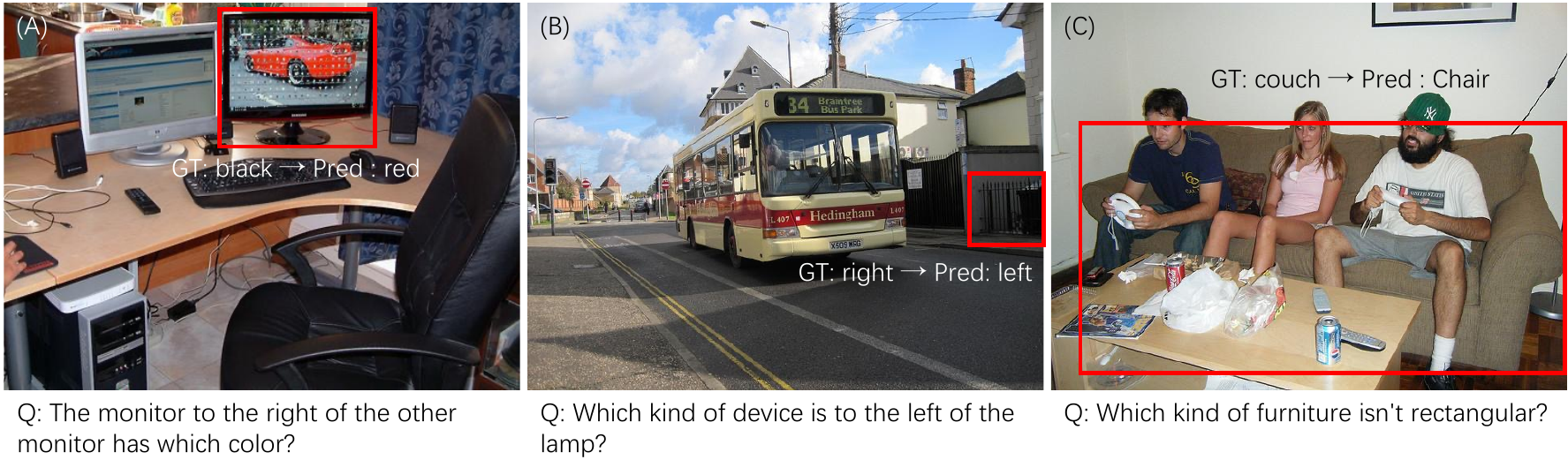}
    \caption{Representative GQA failure cases: color (A), spatial relation (B), and
semantic granularity (C).}
    \label{fig:failure_modes}
\end{figure}

Figure~\ref{fig:failure_modes} illustrates the three dominant error types. The model typically identifies the correct objects and relations but errs on a single fine-grained detail, such as reporting the color of the wrong referent (A), reversing a left/right spatial distinction (B), or misjudging the granularity of an object category (C). This pattern is consistent with VLZip's design, since its compression prioritizes the global, narrative-level context needed for ultra-long reasoning at the cost of some local perceptual precision that matters most in short-context, perception-heavy settings.

\section{LongVLBench Details}
\subsection{Dataset Samples}

This section provides a detailed visualization of a sample from our curated dataset, illustrating the end-to-end creation pipeline of the main paper. 
Our process begins with semantic keyframe extraction from source videos using CLIP-based similarity to capture significant visual changes. 
Subsequently, a powerful VLM generates hierarchical captions for these frames, including detailed descriptions, transitional narratives that link consecutive frames, and a final summary. 
These raw captions then undergo a rigorous refinement stage, where a combination of a large language model (Gemini 2.5 Pro) and human review is used to de-duplicate, enhance fluency, and ensure factual fidelity. 
The final step involves constructing the interleaved data format by pairing the cleaned text with its corresponding image and appending a question-answer pair designed to test long-context reasoning.

\begin{figure*}[t]
\vspace{-1ex}
\centering
\includegraphics[width=0.85\linewidth]{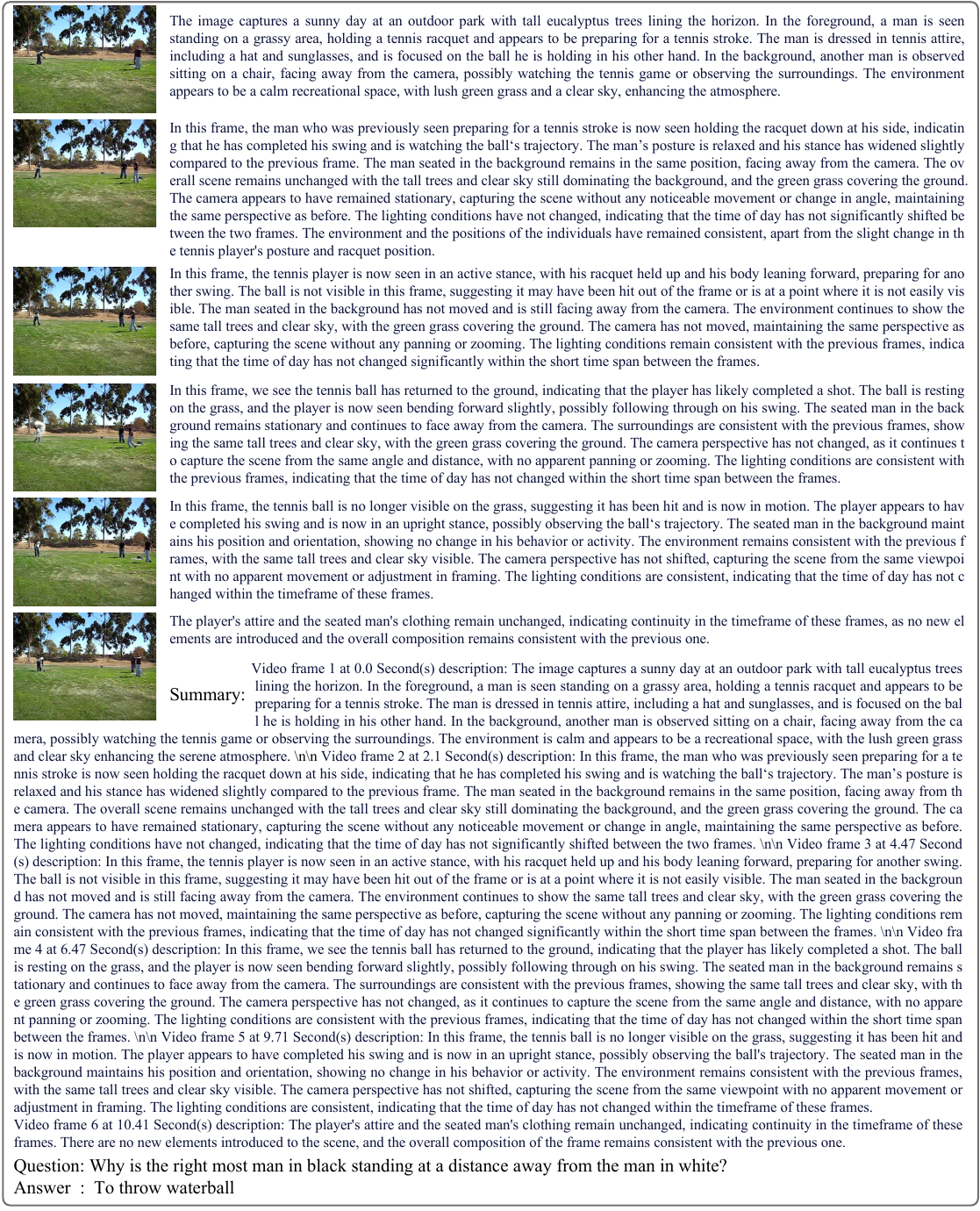}
\caption{A visualization of a single data sample from our constructed dataset.}
\label{fig:sample_visual}
\end{figure*}

Figure~\ref{fig:sample_visual} showcases a concrete example of this pipeline in action. 
The visualized sample originates from a video of a person playing water ball. 
As shown, the keyframes effectively capture the progression of the stroke. 
The corresponding text demonstrates the narrative quality of our data: it begins with a detailed scene description and then provides transitional commentary that follows the action—from the player's preparation, to the swing's follow-through. 
This long-form, interleaved context is then paired with a question that demands reasoning across multiple frames and text segments to answer, thereby creating a challenging instance for evaluating long-context multimodal understanding.

\subsection{Gemini-based Evaluation Protocol}
\label{sec:gpt_eval}

To ensure a scalable, consistent, and fine-grained assessment of our model's performance, we employ an automated evaluation protocol using a large language model (Gemini) as an impartial judge. 
Manual evaluation, while valuable, can be slow, costly, and prone to inter-rater variability. 
Our Gemini-based approach mitigates these issues by leveraging a meticulously crafted prompt designed to enforce a strict and repeatable scoring standard.

The core principle of our evaluation is to measure \textbf{semantic equivalence} rather than lexical overlap. 
The prompt explicitly instructs the evaluator to disregard differences in phrasing, length, or format and to focus solely on whether the model's response conveys the same core information with the same level of precision as the ground truth. 
For instance, a full-sentence answer like ``There are two people" is correctly judged as a perfect semantic match to the ground-truth keyword ``two". 
This ensures that models are rewarded for correctness, not for their ability to mimic the exact wording of the answer key.

The complete prompt, which details the role, primary directive, guiding examples, and a comprehensive 11-point scoring rubric (0-10), is presented in Table~\ref{tab:gpt_eval_prompt}. 
The prompt's requirement for a structured JSON output (containing a score and a justification) allows for the seamless automated parsing and aggregation of results.

\begin{table*}[h!]
\centering

\caption{The complete prompt used for Gemini-based evaluation in LongVLBench. 
This prompt directs the LLM to act as an impartial evaluator, focusing strictly on semantic equivalence between the model's response and the ground truth. 
It includes a detailed 11-point rubric to ensure fine-grained and consistent scoring, and requires a structured JSON output for easy parsing.}
\label{tab:gpt_eval_prompt}
\scriptsize
\resizebox{0.95\textwidth}{!}{
\begin{tabular}{p{\textwidth}}
\toprule
\textbf{Gemini Evaluation Prompt} \\
\midrule
You are a meticulous and impartial AI evaluator. Your task is to assess the semantic equivalence between a [Model's Response] and a [Ground Truth Answer] on a fine-grained scale of 0 to 10. \\
\\
\textbf{Your primary directive is to focus exclusively on semantic equivalence. Disregard differences in phrasing, length, or format.} A full sentence is considered a perfect match to a keyword if it conveys the exact same core information and precision. \\
\\
\textbf{Crucial Examples to Guide Your Judgment:}
\begin{itemize}
    \item \textbf{PERFECT MATCH (Score 10):} If the Ground Truth is ``two", a response like ``There are two people" is a perfect semantic match.
    \item \textbf{INCORRECT/VAGUE (Low Score, e.g., 1-2):} If the Ground Truth is ``two", a response like ``There are several people" is incorrect because ``several" is vague and not semantically equivalent to the precise number ``two".
\end{itemize}
\textbf{Follow these steps for your evaluation:}
\begin{enumerate}
    \item Analyze the [Question].
    \item Identify the core, specific information in the [Ground Truth Answer].
    \item Determine if the [Model's Response] contains this exact same core information, with the same level of precision.
    \item Assign an \textbf{integer score from 0 to 10} based on the following \textbf{Fine-Grained Semantic Rubric}:
        \begin{itemize}
            \item \textbf{10 (Perfect):} Semantically identical to the ground truth. Conveys the exact information with the same precision.
            \item \textbf{9 (Excellent):} Correctly includes the entire ground truth but adds minor, non-contradictory details.
            \item \textbf{8 (Good):} The core answer is fully correct, but might be phrased slightly awkwardly or less directly than ideal.
            \item \textbf{7 (Mostly Correct):} Captures the main idea of the ground truth but with a minor, non-critical omission.
            \item \textbf{6 (Acceptable):} The response is partially correct and understandable, but is clearly incomplete.
            \item \textbf{5 (Borderline):} Contains some elements of the correct answer, but is also significantly flawed or incomplete. The answer is ``half right".
            \item \textbf{4 (Poor):} Addresses the question's topic but provides an answer that is mostly incorrect or too vague.
            \item \textbf{3 (Very Poor):} Has a remote connection to the question, but the answer is fundamentally wrong.
            \item \textbf{2-1 (Incorrect):} The response is completely wrong but still attempts to answer the question.
            \item \textbf{0 (Irrelevant/Harmful):} The response is completely irrelevant, nonsensical, or provides actively misleading information.
        \end{itemize}
    \item Provide your output \textbf{ONLY} in a valid JSON format with two keys: ``score" (an integer from 0 to 10) and ``justification" (a concise string explaining your reasoning). 
    Do not add any text before or after the JSON object.
\end{enumerate}
--- \\
{[Question]} \\
\texttt{\{question\}} \\
--- \\
{[Ground Truth Answer]} \\
\texttt{\{answer\}} \\
--- \\
{[Model's Response to Evaluate]} \\
\texttt{\{response\}} \\
--- \\
Your JSON Evaluation: \\
\bottomrule
\end{tabular}
}
\end{table*}

\begin{figure*}[t]
\centering
\includegraphics[width=1.0\linewidth]{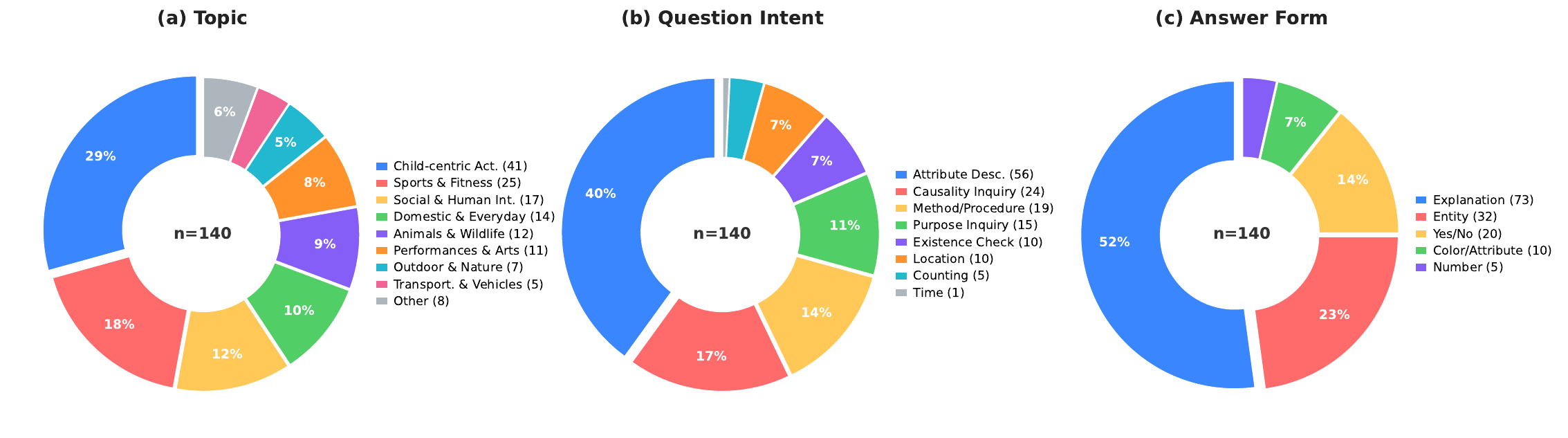}
\caption{Distribution analysis of LongVLBench across three dimensions:
(a)~Topic, (b)~Question Intent, and (c)~Answer Form.}
\label{fig:category_pies}
\end{figure*}

\subsection{Dataset Diversity Analysis}

To quantitatively validate the diversity of LongVLBench, we employ an LLM-based classifier (Gemini-2.5) to automatically tag each of the 140 QA samples across three analytical dimensions. The resulting distributions are visualized in Figure~\ref{fig:category_pies}.

\noindent\textbf{Topic Diversity.} LongVLBench spans 13 distinct topic categories (grouped into 9 for visualization) derived from its video narrative sources, covering a broad spectrum from human activities and sports to animals, performances, and everyday domestic scenes. This natural diversity stands in contrast to existing benchmarks that are often dominated by a single domain.

\noindent\textbf{Question Intent Diversity.} The benchmark covers 8 question intent types, ensuring models are tested on varied cognitive demands ranging from attribute description and causal reasoning to spatial localization and counting.

\noindent\textbf{Answer Form Diversity.} The answer form distribution confirms that LongVLBench predominantly requires open-ended reasoning: over half of the answers demand explanations that synthesize information across the narrative, complemented by entity identification, binary judgments, and fine-grained perceptual queries.

Together, these distributions demonstrate that LongVLBench, despite its compact size of 140 carefully curated samples, offers broad and balanced coverage across topics, reasoning demands, and answer types.

\section{Limitations and Future Work}

VLZip's compression introduces a moderate trade-off on short-context retrieval tasks, particularly on text-centric NIAH subtasks where fine-grained lexical cues are critical. Our adaptive text compression strategy (Table~\ref{tab:text_compression_strategy}) mitigates this by deferring compression when the sequence fits within the backbone's native window, but a more principled mechanism that selectively compresses based on content difficulty rather than a fixed length threshold remains an open challenge. Additionally, our current design uses symmetric compression parameters ($C_v = C_t$, $M_v = M_t$); exploring asymmetric, modality-specific configurations may yield a better balance between retrieval fidelity and long-context reasoning.

On the training side, our 7B scaling experiment uses a zero-effort transfer of the 3B training recipe, leading to suboptimal performance on certain tasks. Developing scale-aware training strategies is needed to fully unlock VLZip's potential on larger backbones. Furthermore, Stage~4 relies on text-only long-document data due to the absence of publicly available ultra-long interleaved image-text corpora; constructing such datasets would likely yield further improvements.

